\documentclass[review]{elsarticle}
\usepackage[margin=2.55cm]{geometry}

\usepackage{lineno,hyperref}
\usepackage[utf8]{inputenc}
\usepackage[T1]{fontenc}
\usepackage{amsmath}
\usepackage{float}
\usepackage{subfig}
\usepackage{caption}

\modulolinenumbers[5]

\usepackage{algpseudocode}
\usepackage{algorithm}

\algnewcommand\algorithmicforeach{\textbf{for each}}
\algdef{S}[FOR]{ForEach}[1]{\algorithmicforeach\ #1\ \algorithmicdo}

\journal{Journal of \LaTeX\ Templates}

\begin{document}

\begin{frontmatter}

\title{A smartphone application to detection and classification of coffee leaf miner and coffee leaf rust}

\author[endufes]{Giuliano L. Manso}
%\address{Universidade Federal do Espírito Santo}

%\author[endufes]{Giuliano L. Manso}

\author[endufes]{Helder Knidel}

\author[endufes,endppgi,endped]{Renato A. Krohling}

\author[endincaper]{José A. Ventura}

\address[endufes]{Labcin - Laboratory of Computing and Engineering Inspired by Nature, UFES - Federal University of Espirito Santo, Vitória, Brazil}

\address[endppgi]{PPGI - Graduate Program in Computer Science, UFES, Vitória, Brazil}

\address[endped]{Production Engineering Department, UFES, Vitória, Brazil}

\address[endincaper]{Incaper, Rua Afonso Sarlo, 160, Bento Ferreira, 29052-010 Vitória, ES, Brazil}


\begin{abstract}
    Generally, the identification and classification of plant diseases and/or  pests are performed  by an expert . One of the problems facing coffee farmers in Brazil is  crop infestation, particularly by leaf rust \textit{Hemileia vastatrix} and leaf miner \textit{Leucoptera coffeella}. The progression of the diseases and or pests occurs spatially and temporarily. So, it is very important to automatically identify the degree of severity. The main goal of this article consists on the development of a method and its  i implementation as an App that allow the detection of the foliar damages from images of coffee leaf that are captured using a smartphone, and identify whether it is rust or leaf miner, and in turn the calculation of its severity degree. The method consists of identifying a leaf from the image and separates it from the background with the use of a segmentation algorithm. In the segmentation process, various types of backgrounds for the image using the HSV and YCbCr color spaces are tested. In the segmentation of foliar damages, the Otsu algorithm and the iterative threshold algorithm, in the YCgCr color space, have been used and compared to \textit{k-means}. Next, features of the segmented foliar damages are calculated. For the classification, artificial neural network trained with extreme learning machine have been used. The results obtained shows the feasibility and effectiveness of the approach to identify and classify foliar damages,  and  the automatic calculation of the severity. The results obtained are very promising according to experts.\\
\end{abstract}

\begin{keyword}
    Segmentation, Feature extraction, Artificial neural networks, Extreme learning machine, Coffee leaf Rust, Coffee leaf miner, \textit{Hemileia vastatrix}, \textit{Leucoptera coffeella}.
\end{keyword}

\end{frontmatter}

%\linenumbers

\section{Introduction}

    Plant diseases \cite{ventura2017} are usually caused by microorganisms, such as bacteria, fungi, nematodes and viruses, but may still be caused by lack or excess of: 1) essentials nutrients for the growth of plants, 2) water and 3) light. In this case, they are also known as physiological disorders. There are many measures that can be taken to avoid the occurrence of disease or even reduce its impact. The observation of a set of control measures and early identification reduce the chance of economic losses with the production and also the use of chemicals products. So, the automatic identification of the diseases and pests are desirable to the health of a plantation. Diseases can affect plants to varying degrees, from small damages until the annihilation of the plantation as a whole. Coffee leaf rust caused by the fungus \textit{Hemileia vastatrix}, is one of the main coffee disease. This disease causes the early fall of the leaves and the consequent drought of the productive branches \cite{garccon2004controle,ventura2017}. When the leaves have a small number of rust foliar damages, the injuries may remain in the plant. However when the severity (percentage of the leaf affected by disease) is high, causes its early fall. In plants susceptible to rust, a single foliar damage may cause the leaf fall \cite{ventura2017}.
    
    The insect \textit{Leucoptera coffeella} (leaf miner) causes reduction in the leaf area and leaf fall with consequent decrease in photosynthesis, resulting in the drop of the production. The larvae of this pest are small that penetrate directly into the leaf mesophyll without touching the outside environment. The damaged regions dry up and the area under attack increases with the development of caterpillars and the various "mines" \cite{fornazier2007pragas}. There are some studies related to the classification of foliar damages on leaves of plants and also in the calculation of severity. Zhang, et al. \cite{zhang2017leaf} proposed an approach to recognition of diseases in cucumber leaf. The method is divided into three main steps: area segmentation to separate leaf damage area using the \textit{k-means} algorithm, extraction of features of the foliar damaged area, and classification using a sparse representation. The great advantage of this approach is the improvement in the recognition performance of diseases and pests. Pahikkala et al. \cite{pahikkala2015classification} presented a study focused on images of overlapping leaves analyzing color photographs of different cultures. The identification of species was based on different textures of monocotyledons and dicotyledons leaves. An automatic classifier based on the learning algorithm Regularized Least-Squares was used.
    
    Aakif, and Khan \cite{aakif2015automatic} proposed an algorithm to identify a plant in three stages: preprocessing, where segmentation is performed with excess green and threshold; in the second stage, extraction of morphological features and application of Fourier descriptors. In the third and last stage, the classification was performed using an artificial neural network. Singh and Misra \cite{singh2017detection} presented an algorithm for segmentation of images that is used to detect the damaged area in leaves of plants for later classification of the injuries. The segmentation of the image was done using a genetic algorithm. Hitimana and Gwun \cite{hitimana2014automatic} proposed an automatic method to detect and estimate severity. The image was processed for background removal and the fuzzy c-means algorithm was applied to the channel V of the YUV color space. In this way, the damaged area becomes evident and severity can be estimated by the ratio of pixels of the injured area divided by the total area of the leaf. Patel and Dewangan \cite{patelautomatic} proposed a mechanism to detect diseases in leaves of plants combining \textit{k-means} and artificial neural networks. The image can be converted to the HSI color space. The \textit{k-means} algorithm can be then applied to determine the region affected by the disease. After identifying the damaged areas, features can be extracted and then used for training and classification. Rastogi, Arora, and Sharma \cite{rastogi2015leaf} proposed a methodology divided into two parts: the first one involves image preprocessing, feature extraction and training an artificial neural network. The second phase also involves the acquisition of image, preprocessing, feature extraction, \textit{k-means}, estimation of disease severity and calculation of the severity using fuzzy logic. Prasetyo et al. \cite{prasetyo2017mango} applied threshold segmentation along with the Otsu method. The method can be applied to the H, S and V channels of the HSV color space and Cb and Cr channels of the YCbCr color space for the mango leaf images. Mwebaze and Owomugisha \cite{mwebaze2016machine} presented a scale (1 to 5) to calculate disease incidence and severity, where 1 means a totally healthy plant and 5 a plant with maximum severity. Mohanty, Hughes, and Salathé \cite{mohanty2016using} presented another approach for classification of plant diseases using convolutional neural networks. There were 14 species and 26 types of diseases. The results presented an accuracy of 99.35\%. So, there are some recent studies for segmentation, feature extraction and classification of plant diseases. The early detection of the disease or pest is fundamental to correct control. In addition, determining the severity of the foliar damaged area may help in decision-making process for the farmer. Garçonet al. \cite{garccon2004controle} shows that a disease control system based on the value of the disease severity is efficient because it saves costs with pulverization. Both tasks recognition and estimation of severity, may be difficult for farmers with little experience. Currently, there are no accessible tools that facilitate these tasks. In most cases, these estimates are made visually or through applications that are not accurate.
    
    In this work, we develop a system for automatically classification of foliar damages of coffee leaves and calculation of its severity. In addition, we show valuable results regarding the image capture process, the background choice, the dataset creation, the best segmentation algorithms, the most suitable neural network for classification and finally the calculation of severity. The system receives as input an image captured by a smartphone, then it identifies the leaf of the coffee tree in the image by means of segmentation, identify the foliar damage contained in the leaf and classify them as well as to calculate the percentage of damaged area. In this case, the damages considered are the coffee leaf rust and coffee leaf miner. It is worth mentioning that there are other diseases and pests not considered in this work \cite{ventura2017}. The article is structured as in the following: Section 2 describes algorithms for segmentation, feature extraction and classification of coffee injured leaves and presents an automatic method to calculate the severity. Sections 3 presents the experiments carried out with comparisons and analysis of the results. Section 4 is development of a mobile solution by implementing an App. Section 5 draw conclusions and present directions for future works.

\section{Algorithms for Digital Image Processing of Coffee Leaves}

    The standard steps in digital image processing  consists of: 1) segmentation; 2) feature extraction, 3) classification. In some cases the pre-processing is included, which may or not be required depending on the problem.
    
    \subsection{Color Space}
    
        \subsubsection{The color space YCbCr}
                    
             In the YCbCr color space, the Y component contains only  luminance. The components of blue  chrominance (B-Y) abbreviated by Cb and red chrominance (R-Y) abbreviated by Cr are not influenced by luminance \cite{chaves2010detecting}. The transformation of RGB to YCbCr is obtained by Equation \ref{eq:rgb2ycbcr} according \cite{chaves2010detecting}.
        
            \begin{equation}
        		\begin{aligned}
            		\resizebox{0.6\hsize}{!}{$%
                        \begin{bmatrix}
                            Y \\
                            Cb \\
                            Cr 
                        \end{bmatrix} = 
                        \begin{bmatrix}
                            16 \\
                            128 \\
                            128 
                        \end{bmatrix} + \frac{1}{256}
                        \begin{bmatrix}
                            65.481 & 128.553 & 24.966 \\
                            -37.797 & -74.203 & 112 \\
                            112 & -93.768 & -18.214 
                        \end{bmatrix} \cdot
                        \begin{bmatrix}
                            R \\
                            G \\
                            B 
                        \end{bmatrix}
                        $%
                        }%
            		\end{aligned}
        		\label{eq:rgb2ycbcr}
            \end{equation}\\
            \noindent where $R$, $G$ and $B$ are the values of the red, green and blue components, respectively, which typically range from 0 to 255. Since the components of chrominance do not vary with lighting, they are widely used in segmentation.
        
        \subsubsection{The color space YCgCr}
            
            Very similar to YCbCr, the YCgCr color space also has chrominance components, but differs from YCbCr by replacing the chrominance component of blue with that of green (G-Y) abbreviated by Cg \cite{de2003face}. The transformation of RGB to YCgCr is obtained by
            Equation \ref{eq:rgb2ycgcr} according \cite{de2003face}.

            \begin{equation}
        		\begin{aligned}
            		\resizebox{0.6\hsize}{!}{$%
                        \begin{bmatrix}
                            Y \\
                            Cg \\
                            Cr 
                        \end{bmatrix} = 
                        \begin{bmatrix}
                            16 \\
                            128 \\
                            128 
                        \end{bmatrix} + \frac{1}{256}
                        \begin{bmatrix}
                            65.481 & 128.553 & 24.966 \\
                            -81.085 & 112 & -30.915 \\
                            112 & -93.768 & -18.214 
                        \end{bmatrix} \cdot
                        \begin{bmatrix}
                            R \\
                            G \\
                            B 
                        \end{bmatrix}
                        $%
                        }%
            		\end{aligned}
        		\label{eq:rgb2ycgcr}
            \end{equation}\\
            \noindent where $R$, $G$ and $B$ are the values of the red, green and blue components, respectively, which typically range from 0 to 255. Since the components of chrominance do not vary with lighting, they are widely used in segmentations.

        \subsubsection{The color space HSV}
        
            The HSV color space represents the colors in terms of matrix or color depth (Hue), abbreviated by H, color purity (Saturation), abbreviated by S, and intensity of the value or brightness of the color (Value), abbreviated by V. The components of the HSV color space is obtained by
            Equations \ref{eq:h}, \ref{eq:s} and \ref{eq:v} according \cite{shaik2015comparative}.
            
            \begin{equation}
            	\begin{aligned}
            	    H = arcos\frac{\frac{1}{2}(2R-G-B)}{\sqrt{(R-G)^2-(R-B)(G-B)}}
            		\end{aligned}
            	\label{eq:h}
            \end{equation}
            
            \begin{equation}
            	\begin{aligned}
            	    S = \frac{\max(R,G,B)-\min(R,G,B)}{\max(R,G,B)}
            		\end{aligned}
            	\label{eq:s}
            \end{equation}
            
            \begin{equation}
            	\begin{aligned}
            	    V = \max(R,G,B)
            		\end{aligned}
            	\label{eq:v}
            \end{equation}
            
            \noindent where $R$, $G$ and $B$ are the values of the red, green and blue components respectively, which typically range from 0 to 255.
    
    \subsection{Segmentation}
    
        The process of segmentation may be thought of as the process of grouping an image in homogeneous units with respect to one or more features [16]. The segmentation of colored images is divided into four groups: edge detection methods; neighborhood-based methods; methods based on histogram; and clustering-based segmentation \cite{lambert2000filtering}.
        
        \subsubsection{Otsu Method}
        
            Segmentation of images based on histograms consists mainly in determining a threshold value. Due to the gray levels that characterize the objects in a grayscale image, it is sought to highlight background objects based on one-dimensional statistics, e.g., histograms of gray levels. Ideally, algorithms do this automatically by selecting the best threshold to determine what is background and what is not (object of interest) of an image \cite{zhang2008image,liu2009otsu}.
            
            The Otsu method \cite{otsu1979threshold} selects a global optimal threshold by maximizing the variance between classes. In a two-level thresholding, the pixel that has a gray level lower than the threshold will be assigned to the background, otherwise it will be considered as an object part \cite{zhang2008image}. 
            
            Assuming that a two-dimensional image is represented in $L$ gray levels $[0,1,...,L-1]$, the number of pixels at level $i$ is denoted by $n_{i}$, and the total number of pixels is denoted by $N = n_{1}+n_{2}+\dots +n_{L}$. Given this distribution, the probability of a gray level is given by Equation \ref{eq:OtsuProb} according \cite{liu2009otsu}.
            
            \begin{equation}
                \begin{aligned}
                	p_{i} = n_{i} / N, ~~p_{i} \ge 0, ~~\sum\limits_{0}^{L-1} p_{i} = 1
                \end{aligned}
                \label{eq:OtsuProb}
            \end{equation}
            
            In a two-class thresholding, the pixels of the image are divided into classes $C_{1}$ with the gray levels $[0,1,...t]$, and $C_{2}$ with the levels $[t+1,...,L-1]$ by the threshold $t$. The probability distributions from the grayscale to the two classes were given by Equantions \ref{eq:OtsuDistProb1} and \ref{eq:OtsuDistProb2} according \cite{liu2009otsu}.
    	    
    	    \begin{equation}
        		\begin{aligned}
            		  \textsc{w}_{1} = \textsc{P}r(C_{1}) = \sum\limits_{i=0}^{t} p_{i}
            	\end{aligned}
        		\label{eq:OtsuDistProb1}
            \end{equation}
            
            \begin{equation}
        		\begin{aligned}
            		\textsc{w}_{2} = \textsc{P}r(C_{2}) = \sum\limits_{i=t+1}^{L-1} p_{i}
            	\end{aligned}
        		\label{eq:OtsuDistProb2}
            \end{equation}
            
            \noindent which is the sum of the probabilities of the gray levels of each one of the classes. The mean of classes $C_{1}$ and $C_{2}$ were respectively, given by Equations \ref{eq:OtsuDistMedia1} and \ref{eq:OtsuDistMedia2} according \cite{liu2009otsu}.
            
            \begin{equation}
        		\begin{aligned}
            	    u_{1} = \sum\limits_{i=0}^{t} ip_{i}/\textsc{w}_{1}
                \end{aligned}
        		\label{eq:OtsuDistMedia1}
            \end{equation}
            
            \begin{equation}
        		\begin{aligned}
            		\resizebox{0.19\hsize}{!}{$%
                        u_{2} = \sum\limits_{i=t+1}^{L-1} ip_{i}/\textsc{w}_{2}
                        $%
                        }%
            		\end{aligned}
        		\label{eq:OtsuDistMedia2}
            \end{equation}
            
            The total mean of the gray levels, represented by $u_{t}$ is calculated by Equation \ref{eq:OtsuDistMediaT} according \cite{liu2009otsu}.
            
            \begin{equation}
        		\begin{aligned}
            		u_{T} = \textsc{w}_{1}u_{1} + \textsc{w}_{2}u_{2}
            	\end{aligned}
        		\label{eq:OtsuDistMediaT}
            \end{equation}
            
            The variances of $C_{1}$ and $C_{2}$ were, respectively calculated by Equations \ref{eq:OtsuDistVar1} and \ref{eq:OtsuDistVar2} according \cite{liu2009otsu}.
            
            \begin{equation}
        		\begin{aligned}
                    \sigma_{1}^{2} = \sum\limits_{i=0}^{t}(i-u_{1})^2p_{i}/\textsc{w}_{1}
            	\end{aligned}
        		\label{eq:OtsuDistVar1}
            \end{equation}

            \begin{equation}
        		\begin{aligned}
            	    \sigma_{2}^{2} = \sum\limits_{i=0}^{t}(i-u_{1})^2p_{i}/\textsc{w}_{2}
            	\end{aligned}
        		\label{eq:OtsuDistVar2}
            \end{equation}
            
            The interclass variance and variance between classes were respectively, calculated by Equations \ref{eq:OtsuDistIC} and \ref{eq:OtsuDistEC} according \cite{liu2009otsu}.
            
            \begin{equation}
        		\begin{aligned}
            		\sigma_{\textsc{w}}^{2} = \sum\limits_{k=1}^{2}\textsc{w}_{k}\sigma_{k}^{2}
            	\end{aligned}
        		\label{eq:OtsuDistIC}
            \end{equation}
            
            \begin{equation}
        		\begin{aligned}
        		    \sigma_{B}^{2} = \textsc{w}_{1}(u_{1}-u_{T})^2 + \textsc{w}_{2}(u_{2}-u_{T})^2
            	\end{aligned}
        		\label{eq:OtsuDistEC}
            \end{equation}
            
            The Otsu method \cite{liu2009otsu} chooses the best threshold $t$ by maximizing variance between classes, which is equivalent to minimizing interclass variance. The threshold value $t$ were calculated by Equations \ref{eq:OtsuT1} and \ref{eq:OtsuT2} according \cite{liu2009otsu}.
            \begin{equation}
        		\begin{aligned}
            		t = arg\{ \max\limits_{0\le t \le L-1} \{ \sigma_{B}^{2} (t) \} \}
                \end{aligned}
        		\label{eq:OtsuT1}
            \end{equation}
            
            \begin{equation}
        		\begin{aligned}
            	    t = arg\{ \min\limits_{0\le t \le L-1} \{ \sigma_{\textsc{W}}^{2} (t) \} \}
                \end{aligned}
        		\label{eq:OtsuT2}
            \end{equation}
            
            The goal is then to iterate through all possible values for the threshold in an image, seeking the one that maximizes the variance between classes. The pseudocode of the Otsu method is described in Algorithm 1.
            
            \begin{algorithm}[H]
                \label{alg:alg1}
                \begin{algorithmic}[1]
                    \Require Image in \textit{L} grey level; 
                    \State $t \gets 0$;
                    \State $t_{max} \gets -1$;
                    \State $variance_{max} \gets -1$;
                    \ForEach {gray level $t$ in $[1,2,...L-1]$}
                        \State $W_{1} \gets Pr(C_{1})$ according to \autoref{eq:OtsuDistProb1};
                        \State $W_{2} \gets Pr(C_{2})$ according to \autoref{eq:OtsuDistProb2};
                        \State Compute mean of classes $C_{1}$ and $C_{2}$ according to \autoref{eq:OtsuDistMedia1} and \ref{eq:OtsuDistMedia2};
                        \State $v \gets \sigma_{B}^2$
                        \If{$v > variance_{max} $}
                            \State $variance_{max} \gets v$;
                            \State $t_{max} \gets t$;
                        \EndIf
                    \EndFor
                    
                    \ForEach {pixel $p$ in the image}
                        \If{$p \ge t_{max} $}
                            \State Define pixel $p$ as white;
                        \EndIf
                        \If{$p < t_{max} $}
                            \State Define pixel $p$ as black;
                        \EndIf
                    \EndFor
                    \State \Return Binary image;
                \end{algorithmic}
                \caption{Otsu Method}
            \end{algorithm}
            
             \subsubsection{\textit{k-means} algorithm}
        
           \textit{k-means} clustering was a method proposed by Macqueen \cite{macqueen1967some} that was commonly used to partition a set of data into $k$ groups \cite{wagstaff2001constrained, liu2009otsu}. 
            
            The algorithm consists in initially selecting $k$ centers of random groups (centroids) in a data set (instances) and iteratively refine them.
            For each instance of the dataset, one calculates the Euclidean distance of this data for each of the centroids by:
            
            \begin{equation}
            	\begin{aligned}
                    dist(\textbf{d},\textbf{c}) = \sqrt{\sum\limits_{i=1}^{n}(d_{i}-c_{i})^{2}}
            	\end{aligned}
            	\label{eq:KmeansEuclidian}
            \end{equation}
            
            \noindent where $\textbf{d} = (d_{1},d_{2},...,d_{n})$ and $\textbf{c} = (c_{1},c_{2},...,c_{n})$ are two points in \textit{n}-dimensional Euclidean space. At each new iteration, the centroid of each group is recalculated and the distances are recalculated again. The algorithm converges when there are no significant changes in the centroids.
            
            For the calculation of each centroid $c_{i}$ of each group $i \in [1,2,...,k]$ is used the mean as described by:
            
            \begin{equation}
            	\begin{aligned}
            		{c}_{i} = \frac{\sum\limits_{j=1}^{n}d_{j}}{n}
            	\end{aligned}
            	\label{eq:KmeansMedia}
            \end{equation}
            
            \noindent where $n$ is the number of data in a given group.
            
            In digital image processing, the input set is all pixels of the image. The distance is calculated with the numerical values of each pixel, i.e., the \textit{k-means} group pixels with similar colors in the same group, thus separating objects belonging to the image.

        \subsubsection{Segmentation on the color space YCgCr}
        
        Feng and He \cite{feng2017color} proposed a threshold-based segmentation method for foliar damages in leaves of plants. The goal is to highlight the injured leaf to extract features of the foliar damages. The method applies iterative threshold segmentation in the color space YCgCr \cite{de2003face}.
                
        The segmentation method initially calculates the difference matrix of \textit{Cr} and \textit{Cg} components according to:
        \begin{equation}
    		\begin{aligned}
    			im_{dif}(x,y) = Cr(x,y) - Cg(x,y)
    		\end{aligned}
    		\label{eq:imDif}
        \end{equation}
        
        \noindent where $im_{dif}$ represents the difference between the pixels of the \textit{Cr} component at the $(x,y)$ position and the \textit{Cg} component at the $(x,y)$ position. The pixels are then separated into two groups, one being the damaged foliar area and the other not, according to the threshold. At each new iteration, the threshold is calculated based on the mean of the two groups and the global mean.
        
        According to \cite{feng2017color} the algorithm works well for segmentation of foliar damages with brown, red and yellow tones. Algorithm 2 presents the pseudocode of iterative thresholding segmentation.
      
            \newpage
            %\begin{figure}[H]
            %\centering
            %    \includegraphics[scale=0.4]{figs/6.png}
            %\caption{\label{fig:sisproc} Subsistemas de processamento de imagem.}
            %\end{figure}
            
        \begin{algorithm}[H]
            \label{alg:alg2}
            \begin{algorithmic}[1]
                \Require \textit{Cg} and \textit{Cr} channels with dimensions \textit{m} x \textit{n}; 
                \State Initializes $T \gets 0$;
                \Repeat
                    \State Divides the $im_{dif}$ matrix using $T$ into two groups;
                    \State $G_{1} \gets im_{dif} > T$;
                    \State $G_{2} \gets im_{dif} \leq T$;
                    \State $m_1 = \frac{\sum\limits_{i=0}^{m-1}\sum\limits_{i=0}^{n-1}G_1(m,n)}{m\times n}$;
                    \State $m_2 = \frac{\sum\limits_{i=0}^{m-1}\sum\limits_{i=0}^{n-1}G_2(m,n)}{m\times n}$;
                    \State The new threshold:
                    \State $T \leftarrow (m_1+m_2)/2$;
                \Until Convergence;
                \State \Return Matrix $G_{1}$ and $G_{2}$;
            \end{algorithmic}
            \caption{Segmentation on the color space YCgCr}
        \end{algorithm}
        
        %\begin{figure}[H]
        %\centering
        %    \includegraphics[scale=0.4]{figs/8.png}
        %\caption{\label{fig:sisproc} Subsistemas de processamento de imagem.}
        %\end{figure}

\subsection{Feature extraction}

    In order to obtain good results in the subsequent stages of processing, it is necessary to create a set of attributes that will be used in the training stage, and later in the classification phase \cite{de2005classificaccao}. It is important to note that, in this step, the input is still an image but the output is a set of measures corresponding to that image.
        
    Extracting features (attributes) from an image highlights differences and similarities between objects. Among these features, one can include the brightness of a region, the texture, the amplitude of the histogram, among others. In general, the extraction of features is a process usually associated to the analysis of the regions of an image \cite{deextraccao}.
    
    \subsubsection{Texture Attributes}
    
        Texture contain important information about the structural arrangement of surfaces and their relationships with the medium \cite{haralick1973textural}. It is possible to obtain the texture attributes by means of the Gray Level Co-occurrence Matrix (GLCM) that was initially described by Harlick et al. \cite{haralick1973textural}. This methodology explores the dependence of the gray levels of the texture to assemble the GLCM. This matrix represents the relative frequency with which neighboring pixels occur in the gray-scale image \cite{de2005classificaccao, haralick1973textural}.
        
        Next, some texture attributes that are measured from the co-occurrence matrix \cite{haralick1973textural} are presented. In the following equations, $L$ represents the gray levels that compose the image, $p(i,j)$ represents the relative frequency with which neighboring pixels occur in the image, $i$ and $j$ the gray levels in the GLCM.
        
        \begin{enumerate}
            \item \textbf{Energy or second angular momentum:} Measures the textural uniformity. High values of this attribute indicate that the distribution of the gray level in the image has a uniform distribution \cite{soares1997investigation}.
            \begin{equation}
        		\begin{aligned}
        			Energy = \sum\limits_{i,j=0}^{L-1}p(i,j)^2
        		\end{aligned}
        		\label{eq:energia}
            \end{equation}
                
            \item \textbf{Contrast:} It is the difference between the highest and lowest values of gray of a set of adjacent pixels \cite{soares1997investigation}.
            \begin{equation}
        		\begin{aligned}
            	    Contrast = \sum\limits_{i,j=0}^{L-1}p(i,j)(i-j)^2
            	\end{aligned}
        		\label{eq:contraste}
            \end{equation}
            
            \item \textbf{Homogeneity:} Measures the homogeneity of an image. This attribute assumes higher (larger) values when there only small differences in tones (shades) of gray in the pixel sets. Therefore, homogeneity and contrast are inversely correlated \cite{soares1997investigation}.
            \begin{equation}
        		\begin{aligned}
            		Homogeneity = \sum\limits_{i,j=0}^{L-1}\frac{p(i,j)}{1+(i-j)^2}
            	\end{aligned}
        		\label{eq:homogeneidade}
            \end{equation}
                
            \item \textbf{Dissimilarity:} This measure increases linearly as it moves away from the diagonal, unlike contrast, which increases exponentially.
            \begin{equation}
        		\begin{aligned}
        		    Dissimilarity = \sum\limits_{i,j=0}^{L-1}p(i,j)|i-j|
        		\end{aligned}
        		\label{eq:dissimilaridade}
            \end{equation}
                
            \item \textbf{Correlation:} It is a measure of linear dependence on the image. High correlation values imply a linear relationship between the gray levels of neighboring pixels \cite{soares1997investigation}
            \begin{equation}
        		\begin{aligned}
            		Correlation = \sum\limits_{i,j=0}^{L-1}\frac{ijp(i,j) - \mu_{x}\mu_{y}}{\sigma_{x}\sigma_{y}}
                    \end{aligned}
        		\label{eq:correlacao}
            \end{equation}
                
            \noindent where,
            \begin{equation}
        		\begin{aligned}
            	    \mu_{x} = \sum\limits_{i}^{L-1}ip(i,\ast)
        		\end{aligned}
        		\label{eq:m1}
            \end{equation}
            
            \begin{equation}
        		\begin{aligned}
            		\mu_{y} = \sum\limits_{j}^{L-1}ip(\ast,j)
                \end{aligned}
        		\label{eq:m2}
            \end{equation}
            
            \begin{equation}
        		\begin{aligned}
            		\sigma_{x} = \sqrt{\sum\limits_{i}^{L-1}(i-\mu_{x})^2p(i,\ast)} 
        		\end{aligned}
        		\label{eq:d1}
            \end{equation}
            
            \begin{equation}
        		\begin{aligned}
        		    \sigma_{y} = \sqrt{\sum\limits_{j}^{L-1}(j-\mu_{y})^2p(\ast,j)}    
        		\end{aligned}
        		\label{eq:d2}
            \end{equation}
            
            \begin{equation}
        		\begin{aligned}
            		p(i,\ast) = \sum\limits_{j}^{L-1}p(i,j)    
        		\end{aligned}
        		\label{eq:pi}
            \end{equation}
            
            \begin{equation}
        		\begin{aligned}
        		    p(\ast,j) = \sum\limits_{i}^{L-1}p(i,j)        
        		\end{aligned}
        		\label{eq:pj}
            \end{equation}
                
        \end{enumerate}

        Other variables used to quantify textures are based on first order statistics that are calculated in a subregion \cite{soares1997investigation}. These first-order attributes do not take into account the spatial distribution of gray levels in a region of the image. They can be calculated by the frequency distribution of the gray levels of the pixels of an image \cite{de2005classificaccao}.
            
        Next, $P(i)$ is the relative frequency with which the gray level $i$ occurs in the region, and $M$ is the average of the gray levels of the region \cite{soares1997investigation}. $L$ are all gray levels in the image.
        
        \begin{enumerate}
            \item \textbf{Variance:} It is a measure of statistical dispersion. The higher the variance, the more distant from the mean will be the levels of gray. It is defined by:
            
            \begin{equation}
        		\begin{aligned}
            		Variance = \sum\limits_{i}^{L-1}[i-M]^2P(i)    
        		\end{aligned}
        		\label{eq:var}
            \end{equation}

            \item \textbf{Kurtosis:} It characterizes the flattening of the curve of the probability distribution function. It is defined by:
            
            \begin{equation}
        		\begin{aligned}
        		    Kurtosis = \frac{\sum\limits_{i}^{L-1}(i-M)^4P(i)}{V^2}
        		\end{aligned}
        		\label{eq:curt}
            \end{equation}
            
            \noindent where $V$ is the variance.
            
            \item \textbf{Entropy:} It measures the clutter of an image. When an image does not have a uniform texture the entropy is high \cite{soares1997investigation}. It is defined by:
            
            \begin{equation}
        		\begin{aligned}
        		    Entropy = -\sum\limits_{i}^{L-1}P(i)*logP(i)    
        		\end{aligned}
        		\label{eq:entropy}
            \end{equation}
            
        \end{enumerate}
        
    \subsubsection{Color attributes}
    
        The values of mean, standard deviation and variance can be used in color attributes. Next, in the Equations \ref{eq:mean} and  \ref{eq:deviation}  $I(i,j)$  represents the pixel value of the image at position $(i,j)$. The input image has dimensions $m\times n$ \cite{mandloi2014survey, monsalve2015automatic}.
            
        \begin{enumerate}
            \item \textbf{Mean:} The mean is the sum of the values of all pixels in the image. It can be applied to each of the RGB channels of the input image. The mean is calculated by:
            
            \begin{equation}
        		\begin{aligned}
        		    \mu =\frac{\sum\limits_{i}^{m-1}\sum\limits_{j}^{n-1}I(i,j)}{m\times n}
        		\end{aligned}
        		\label{eq:mean}
            \end{equation}

            \item \textbf{Deviation:} For each RGB channel of the image, the standard deviation is calculated by:
            
            \begin{equation}
        		\begin{aligned}
            		\sigma =\frac{\sum\limits_{i}^{m-1}\sum\limits_{j}^{n-1}(I(i,j)-\mu)^2}{m\times n}
        		\end{aligned}
        		\label{eq:deviation}
            \end{equation}
            
        \end{enumerate}
        
        Methods for extraction of attributes allows a better discrimination of the classes in the classification process. In this work, the attributes are represented by vectors containing values extracted from the images \cite{deextraccao}.\\
        
        For this work, the following attributes were selected after an in depth experimental study:
        
        \begin{itemize}
            \item Distributional attributes: mean, standard deviation, kurtosis and entropy. Each of them is applied to each RGB component of the injured foliar area \cite{monsalve2015automatic}.
            
            \item Attributes of co-occurrence: contrast, dissimilarity, homogeneity, energy and correlation. All attributes are calculated from the co-occurrence matrix \cite{singh2017detection}.
        \end{itemize}

\subsection{Classification Algorithms}

    Classification is the process of assigning a label to an object based on its features translated by its descriptors. Data classification is present in several real problems such as: recognizing patterns in images, differentiating species of plants, classifying between benign and malignant tumors, among others \cite{aggarwal2014data}. There are several algorithms for classifying data. In this work, Artificial Neural Network (ANN) trained with Backpropagation, and Extreme Learning Machine (ELM) are used.
    
    \subsubsection{Artificial Neural Networks}
    
        Artificial Neural Networks can be described as a mapping of a input set to an output set \cite{priddy2005artificial, haykin2009neural}. It resembles human brain in two aspects: 1) Knowledge is acquired by the network through a learning process. 2) Interconnections between neurons, known as synaptic weights, are used to store the knowledge acquired. A neuron is a fundamental processing unit for the operation of a neural network. They are simplifications of the biological neuron \cite{mcculloch1943logical}.
        
        The basic elements in a neuron are \cite{haykin2009neural}:
        a) A set of synapses where each of them has an associated weight. Specifically, a signal $x_j$ at the input of the synapse $j$ connected to neuron $k$, is multiplied by the synaptic weight $w_{kj}$. b) An adder, to sum up all the input signals multiplied by the respective synaptic weights. The operations described here constitute a linear combination. c) An activation function to limit the output amplitude of each neuron. Normally, the variation of the output amplitude $y_k$ of neuron $k$ is a closed interval in the range [0,1] or [-1,1].
        
\begin{figure}[H]
            \centering
            \includegraphics[scale=1]{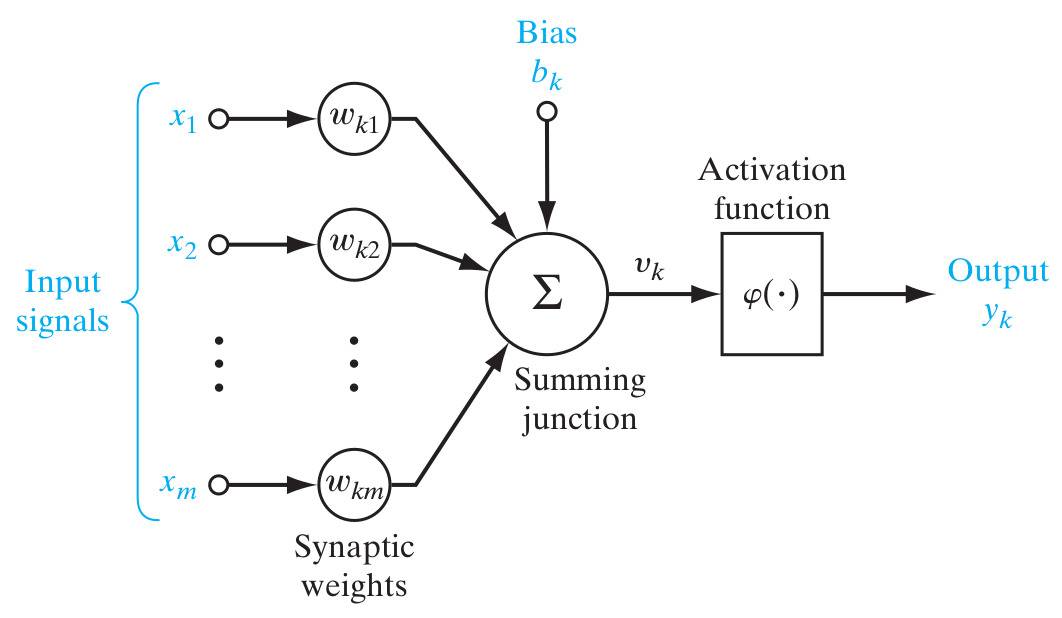}
            \caption{\label{fig:figura1} Representation of an artificial neuron.}
        \end{figure}

        \autoref{fig:figura1} shows the model of an artificial neuron.
        
        In the summing junction, the bias ($b_k$) is also taken into account. Therefore, a neuron $k$ is described by:
            
        \begin{equation}
    		\begin{aligned}
    			y_k = \varphi(\textbf{x},\textbf{w},\textbf{b}) = \varphi(\sum\limits_{j=1}^mw_{kj}x_j+b_k)
    		\end{aligned}
    		\label{eq:neuron}
        \end{equation}
        
        \noindent where $x_j$ are the scalar values of the input, $m$ is the size of the input, $b_k$ are the bias, $w_k$ are the weights learned by ANN and $\varphi(\cdot)$ is the activation function of the neuron. The activation function $\varphi(\cdot)$, defines the output of the neuron in terms of $v_k$. The most frequently used are: the sigmoid, the hyperbolic tangent and Relu \cite{haykin2009neural}.
	    
        An ANN architecture is related to the way the neurons are connected to each other, how they are grouped in layers and the learning algorithms used in training of the weights. The multi-layer feedforward network has 3 types of layers, where the outputs of a layer are the inputs of the following layer. The architecture is made up of: 1) Input layer, which is the layer connected immediately to the values of the inputs to be processed. 2) Intermediate or hidden layers, the neurons corresponding to these layers are called hidden neurons. The term hidden corresponds to the fact that this part of the network can not be seen directly from the input or output. The addition of hidden layers of neurons enables the extraction of several high-order statistics. These layers are responsible for learning the representation of network information \cite{haykin2009neural}. 3) Output layer: In this layer, the goal is to adjust the internal information of the network in a format suitable for use.

\begin{figure}[H]
            \centering
            \includegraphics[scale=0.8]{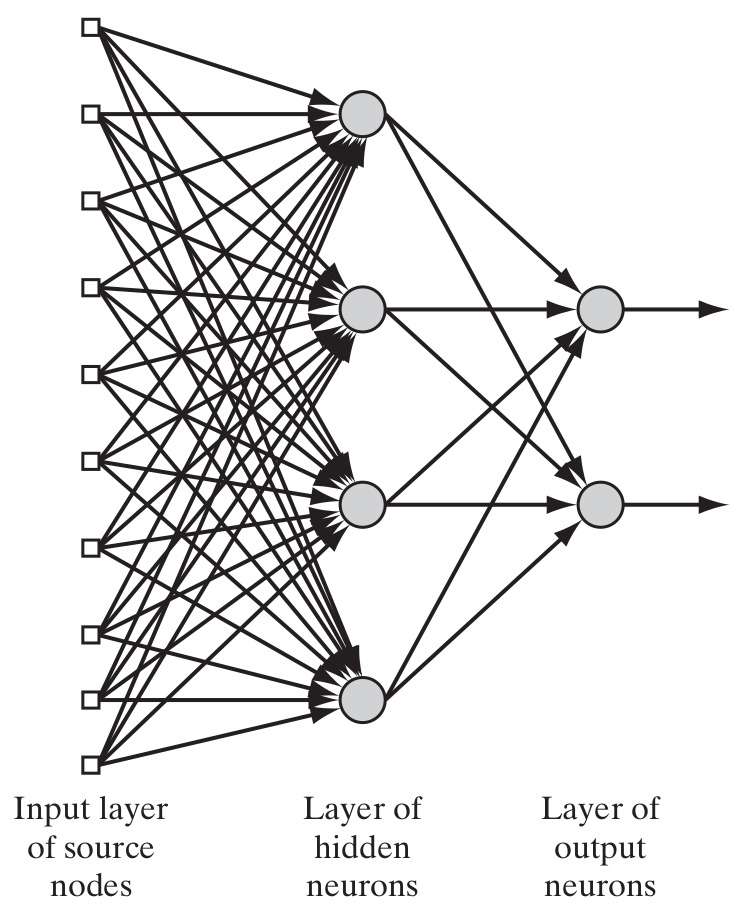}
            \caption{\label{fig:ann} Single Layer Feedforward Neural Network.}
        \end{figure}

        In \autoref{fig:ann}, is shown a Single Feedforward Neural Network (SLFN) with a single hidden layer, 10 input values, 4 neurons in the hidden layer, and 2 neurons in the output layer.

        Learning in a neural network is a process by which the parameters (synapses $\textbf{w}$) of the network are adapted through a process of stimulation by the environment in which the network is inserted. The type of learning is determined by the way in which the modification of the parameters occurs. Two basic learning paradigms are unsupervised and supervised learning \cite{haykin2009neural}. In this work we  use two supervised learning algorithms for training the SLFN: 1) Backpropagation and 2) Extreme Learning Machines (ELM).
        
        \subsubsection{Backpropagation algorithm}
        
        When using supervised learning, it is necessary to minimize a cost function that computes the error between the expected output and the result obtained by the network output. Widely used is the mean square error, which is decribed by:
        
		\begin{equation}
    		\begin{aligned}
    		    E(\textbf{x,w,ŷ}) = \frac{1}{m} \sum\limits_{k=0}^m(\textbf{ŷ}^{(k)}-\varphi(\textbf{x}^{(k)},\textbf{w}))^2
    		\end{aligned}
    		\label{eq:error}
	    \end{equation}
        
       The Gradient descent is described by:
        
        \begin{equation}
    		\begin{aligned}
    		    \textbf{w} = \textbf{w} - \alpha\frac{\partial E(\textbf{x},\textbf{w},\textbf{ŷ})}{\partial\textbf{w}}
    		\end{aligned}
    		\label{eq:gradient}
	    \end{equation}

        \noindent The backpropagation is the algorithm most used to minimize this function. So, the error is moved down successively towards a minimum point of the error surface; whereas the minimum point may be a local or a global minimum \cite{haykin2009neural}.
        
	    In \autoref{eq:gradient} the variable $\alpha$ corresponds to the learning rate. The peseudo-code for the network training is described in Algorithm 3.

        \begin{algorithm}[H]
            \label{alg:alg3}
            \begin{algorithmic}[1]
                \Require Training data set $(\textbf{x},\textbf{ŷ})$;
                Learning rate $\alpha$;
                \State $\textbf{w} \leftarrow$random initialization of the weights according to a uniform distribution in range [-1,1];
                \Repeat
                    \State $a \gets X$;
                    \ForEach {layer $ [\textsc{\textbf{1}},\textsc{\textbf{2}},...| \textsc{\textbf{w}}|]$}
                        \State $a \gets f(a,w_{i})$ according to $\autoref{eq:neuron}$;
                    \EndFor
                    \State Update weights according to \autoref{eq:gradient};
                \Until Convergence;
                \State \Return Matrix $G_{1}$ and $G_{2}$.
            \end{algorithmic}
            \caption{Neural Network Training with Gradient Descent}
        \end{algorithm}
        
        %\begin{figure}[H]
        %    \centering
        %    \includegraphics[scale=0.4]{figs/14.png}
        %\end{figure}

    \subsubsection{Extreme Learning Machine}
    
        Essentially, extreme learning machine was originally developed for SLFN. ELM aims to find not only the smallest training error but also the lower norm of the output weights \cite{huang2012extreme}.
        
        The weights of the input layer neurons and bias are initialized with random values, and the weights of the neurons of the output layer are calculated analytically without using iterative processes. The mathematical formulation of ELM is described by \cite{huang2004extreme, huang2006extreme} through the following equation:
        
        \begin{equation}
        	\begin{aligned}
                \sum\limits_{i=1}^{\tilde{N}} \pmb{\beta}_{i}g(\textbf{w}_{i}\cdot\textbf{x}_{j}+b_{i})=\textbf{t}_{j}, ~~~ j=1,...,N
        	\end{aligned}
        	\label{eq:elm}
        \end{equation}
        
        \noindent where,
        
        \begin{itemize}
            \item $\tilde{N}$ is the number of hidden neurons and $N$  is the number of training samples.
            
            \item $\textbf{x}_{j} = [x_{j1},x_{j2},...,x_{jn}]^{\textbf{T}}$ is the $j$-th input vector and represents each different samples.
            
            \item $\textbf{t}_{j} = [t_{j1},t_{j2},...,t_{jm}]^{\textbf{T}}$ is the $j$-th output target vector and represents the expected outputs with respect to a sample input vector $\textbf{x}_{j}$.
            
            \item $\textbf{w}_{i} = [w_{i1},w_{i2},...,t_{in}]^{\textbf{T}}$ represents the vector of weights that connects the \textit{i}-th neuron of the hidden layer to the neurons of the input layer.
            
            \item $\pmb{\beta}_{i} = [\beta_{i1},\beta_{i2},...,\beta_{im}]^{\textbf{T}}$ represents the vector of weights that connects the \textit{i}-th neuron of the hidden layer to the neurons of the output layer.
            
            \item $b_{i}$ represents the bias associated to the \textit{i}-th neuron of the hidden layer.
            
            \item $g(\cdot)$ is the activation function.
        \end{itemize}

        The $N$ equations presented above are described in simplified form by $\pmb{H}\hat{\pmb\beta} = \pmb{T}$, whose matrix form is given by:
        
        \begin{equation}
        	\begin{aligned}
                \textbf{H} = 
                    \begin{bmatrix}
                        g(\textbf{w}_{1}\cdot\textbf{x}_{1}+b_{1})    & \dots  &    g(\textbf{w}_{\tilde{N}}\cdot\textbf{x}_{1}+b_{\tilde{N}})  \\
                        \vdots & \ddots & 	\vdots \\
                        g(\textbf{w}_{1}\cdot\textbf{x}_{N}+b_{1})    & \dots  &    g(\textbf{w}_{\tilde{N}}\cdot\textbf{x}_{\tilde{N}}+b_{\tilde{N}})
                    \end{bmatrix}_{N\times\tilde{N}}  ~ 
                    \hat{\pmb{\beta}} = \begin{bmatrix} 
                        \pmb{\beta}_{1}^{\textbf{T}} \\ \vdots \\ \pmb{\beta}_{\tilde{N}}^{\textbf{T}}
                    \end{bmatrix}_{\tilde{N}\times m}  ~  
                    \textbf{T} = \begin{bmatrix} 
                        \textbf{t}_{1}^{\textbf{T}} \\ \vdots \\ \textbf{t}_{N}^{\textbf{T}}
                    \end{bmatrix}_{N\times m}
        	\end{aligned}
        	\label{eq:matH}
        \end{equation}
        
        The determination of the output weights, which connect the neurons of the hidden layer to the output layer is defined as the Least-Squares solution of the linear system $\pmb{H}\hat{\pmb\beta} = \pmb{T}$, which is given by $\hat{\pmb\beta} = \pmb{H}^{\dagger}\pmb{T}$, where $\pmb{H}^{\dagger}$ is the generalized inverse of Moore-Penrose's matrix $\pmb{H}$ \cite{huang2004extreme, huang2006extreme}. The pseudo-code of the ELM training is presented in Algorithm 4.
        
        \begin{algorithm}[H]
            \label{alg:alg4}
            \begin{algorithmic}[1]
                \Require Training data set $(\textbf{x},\textbf{ŷ})$;
                Learning rate $\alpha$;
                
                \State Preprocessing of the input samples;
                \State Initialization of weights and bias between the first and hidden layer with random values;
                \State Set the number of neurons in the hidden layer;
                
                \State Calculation of the matrix $H$ according to \autoref{eq:matH};
                \State Calculation of $\hat{\beta}$ according to $\hat{\beta}=H^{\dagger}T$;
                \State \Return $\hat{\beta}$.
            \end{algorithmic}
            \caption{ELM training}
        \end{algorithm}

     \subsection{Severity calculation}
    
        Knowing the presence or absence of disease or pest is important for the farmer. However, knowing the severity of the disease is extremely important so that measures can be taken to avoid loss of crop yields. In addition, this measure brings important information over time on the resistance of the disease or pest and its progress \cite{mwebaze2016machine, ventura2017}.
        
        There are some methods for estimating the damaged foliar area. The two most common are: 1) the estimation of the severity by manual calculation of the damaged foliar area in the leaf with the use of measuring tools; 2) estimation visually based on a diagrammatic scale.
        
        In this work, the severity estimation consists of the count of the image pixels belonging to the injured area of the leaf. One calculates the severity according to
        
        \begin{equation}
        	\begin{aligned}
        	    Severity = \frac{A_{damaged}}{A_{leaf}} \times 100
        	\end{aligned}
        	\label{eq:severity}
        \end{equation}

        \noindent where, $A_{damaged}$ is the area of the injured region of the leaf image in pixels, and the $A_{leaf}$ is the total area of the leaf in pixels.

    \subsection{Proposed Approach}
        
\begin{figure}[H]
            \centering
            \includegraphics[scale=2]{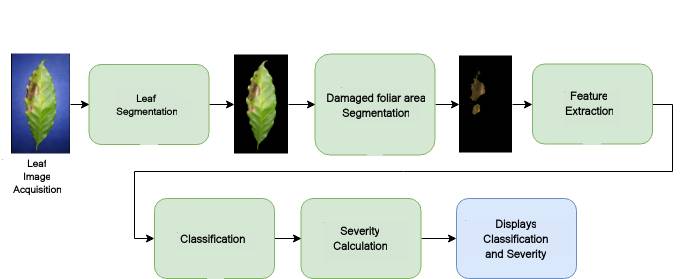}
            \caption{\label{fig:generalSystem} Overview of the automatic process for  classification of the damaged foliar area and calculation of the degree of severity of the coffee leaf from images obtained with smartphone.}
        \end{figure}

        It is expected a system for automatic classification of damaged foliar in coffee leaves and calculation of severity. The system receives as input an image captured by the smartphone, then identify the leaf of the coffee in the image by means of the segmentation, identify the damaged foliar area contained in the leaf and classify. In addition, it calculates the percentage of injured area as shown in \autoref{fig:generalSystem}. In this study, the damaged foliar area considered in the leaf of the coffee tree was the coffee leaf miner and the coffee leaf rust, although there were other diseases and pests not considered in this work \cite{ventura2017}.
        
\section{Experimental results}

    \subsection{Database}
        
        The images used in this work were captured using the ASUS Zenfone 2 smartphone (ZE551ML) with a resolution of 10 Megapixels (4096x2304 pixels). Three background colors were used: white; black and blue. Table 1 shows the amount of images contained in the dataset. The database consists of 690 images, divided according to \autoref{tab:base}. They are available from authors upon request.
        
        \begin{table}[H]
            \centering
            \resizebox{0.8\textwidth}{!}{% <------ Don't forget this %
            \begin{tabular}{|l|c|c|c|c|}
            \hline
            Class       & White background & Blue background & Black background & Total number \\ \hline
            Normal        & 58           & 58         & 58          & 174                   \\ \hline
            Leaf miner & 88           & 88         & 88          & 264                   \\ \hline
            Rust      & 84           & 84         & 84          & 252                   \\ \hline
            \end{tabular}% <------ Don't forget this %
            }
            \caption{Dataset division.}\label{tab:base}
        \end{table}
        
        \begin{figure}[H]
            \centering
            \subfloat[Healthy leaf.\label{fig:l1}]
            {\includegraphics[width=0.2\textwidth]
                {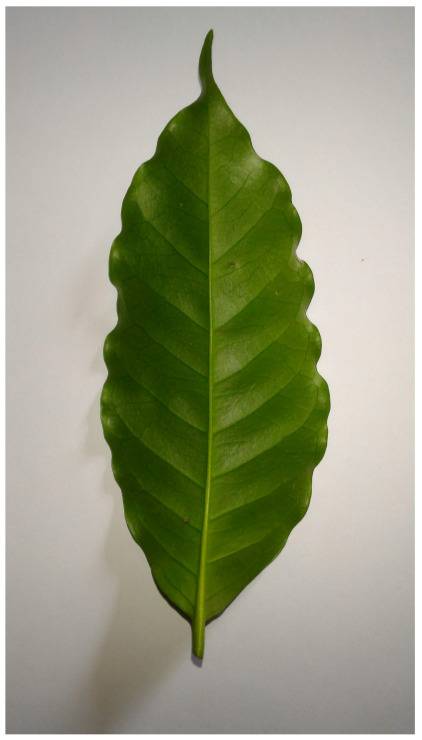}}
            \subfloat[Diseased leaf.\label{fig:l2}]
            {\includegraphics[width=0.2\textwidth]
                {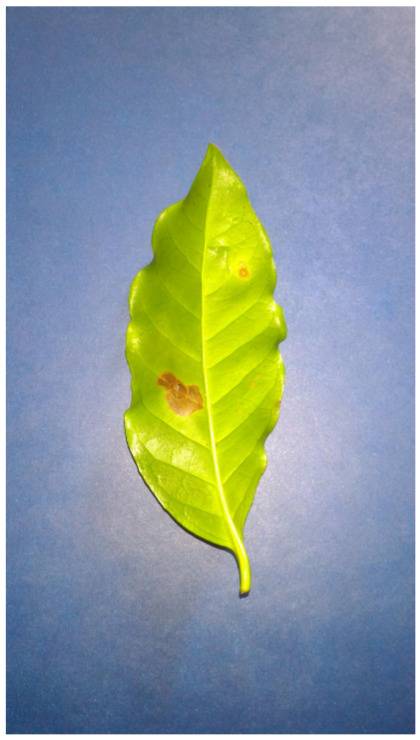}}
            \subfloat[Diseased leaf.\label{fig:l3}]
            {\includegraphics[width=0.2\textwidth]
                {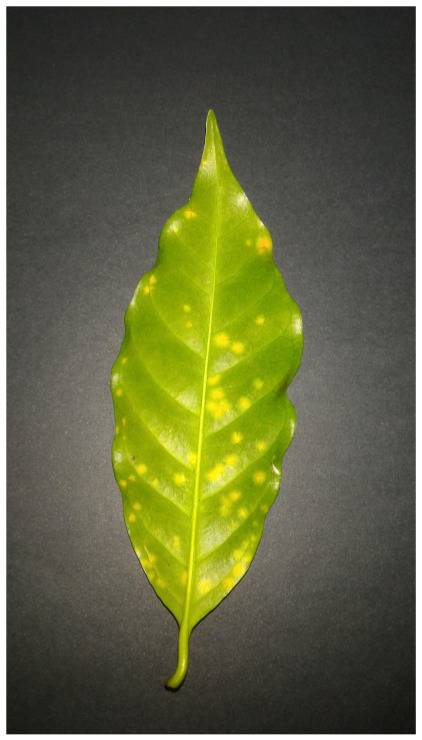}}
            \caption{Images of coffee leaf submitted to the segmentation process.} 
            \label{fig:imagensFundos}
        \end{figure}
        
        The leaves containing coffee leaf miner and coffee leaf rust were segmented from 230 images with white background and divided into two classes according to \autoref{tab:base2}.
    
        \begin{table}[H]
            \centering
            \resizebox{0.3\textwidth}{!}{% <------ Don't forget this %
            \begin{tabular}{|l|c|}
                \hline
                Class       & \multicolumn{1}{l|}{Number of images} \\ \hline
                Coffee leaf miner & 256                                        \\ \hline
                Coffee leaf Rust      & 759                                        \\ \hline
            \end{tabular}% <------ Don't forget this %
            }
            \caption{Imbalanced dataset.}\label{tab:base2}
        \end{table}
    
        For training of the classification algorithms is used the balanced dataset as shown in \autoref{tab:base3}. The class imbalance problem in training of the classifier usually does affect performance, especially for small and moderate training data sets that contain correlated or uncorrelated features \cite{mazurowski2008training}.
        
        \begin{table}[h!]
            \centering
            \resizebox{0.3\textwidth}{!}{% <------ Don't forget this %
            \begin{tabular}{|l|c|}
                \hline
                Class       & \multicolumn{1}{l|}{Number of images} \\ \hline
                Coffee leaf miner & 256                                        \\ \hline
                Coffee leaf rust      & 256                                        \\ \hline
            \end{tabular}% <------ Don't forget this %
            }
            \caption{Balanced dataset.}\label{tab:base3}
        \end{table}

        \begin{figure}[H]
            \centering
            \subfloat[Coffee leaf miner.\label{fig:miner}]
            {\includegraphics[width=0.3\textwidth]
                {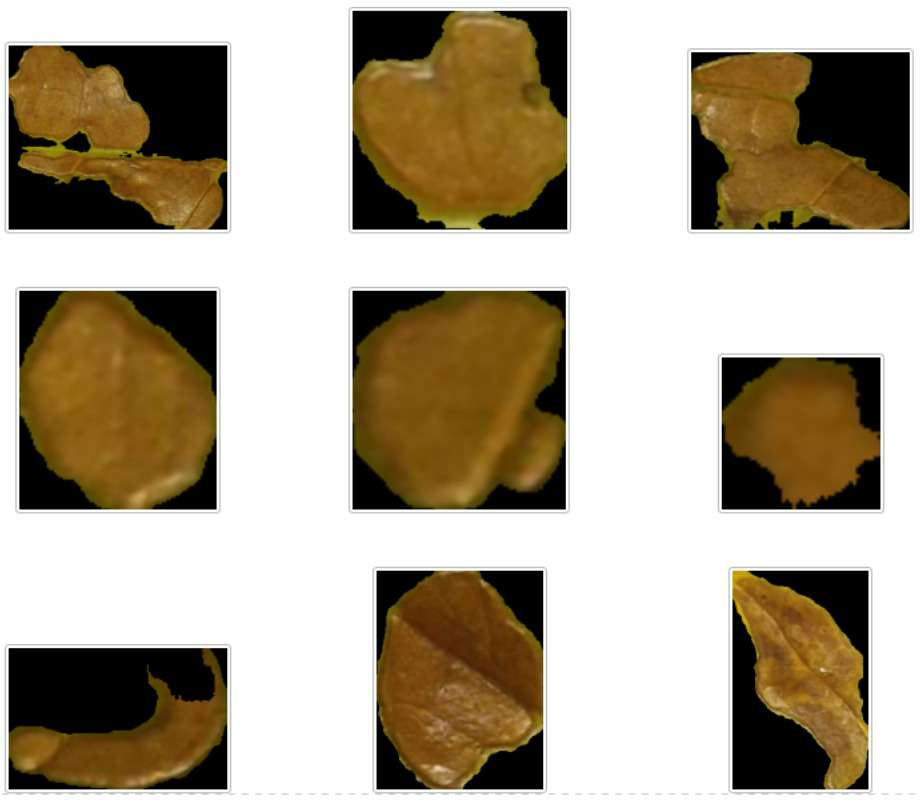}}
            \subfloat[Coffee leaf rust.\label{fig:rust}]
            {\includegraphics[width=0.3\textwidth]
                {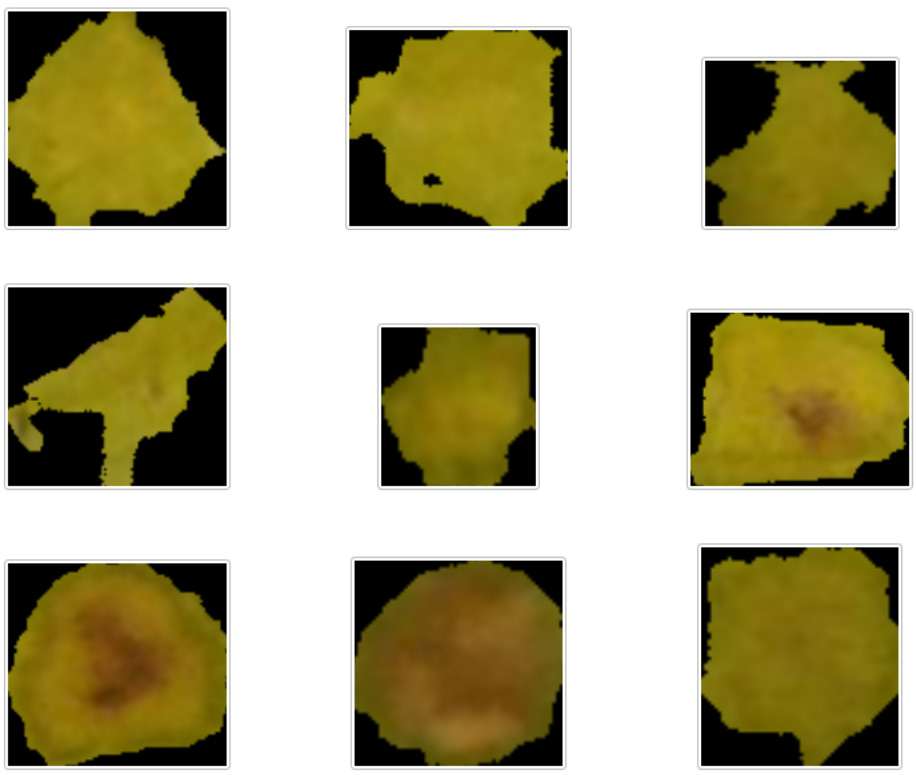}}
            \caption{Foliar damages} 
            \label{fig:lesoesBF}
        \end{figure}
        
        %\begin{figure}[H]
        %    \centering
        %    \includegraphics[scale=0.2]{figs/16.png}
        %    \caption{\label{fig:lesoesBF} Leaf lesions.}
        %\end{figure}
    
    \subsection{Evaluation of the Segmentation Results}
        
        The metric used to evaluate the segmentation algorithms is the same as used in \cite{feng2017color} and \cite{meyer2008verification}. The metric is defined by the Recognition Working Group (Automatic Target Recognition Working Group (ATRWG). The segmentation precision is defined by:
            
        \begin{equation}
    		\begin{aligned}
        		Q_{seg} = \frac{\sum\limits_{k,j=0}^{k,j=m,n}(A(i)_{k,j} \cap B(i)_{k,j})}{\sum\limits_{k,j=0}^{k,j=m,n}(A(i)_{k,j} \cup B(i)_{k,j})}
        	\end{aligned}
    		\label{eq:segQuality}
        \end{equation}
        
        \noindent where $A$ is the segmentation of the algorithm under evaluation, $B$ is the manual segmentation that should be optimal. The index $i=255$ indicates the pixels that are considered as leaf and $i=0$ are the pixels considered as background. $k$ and $j$ are the row and column indices respectively of the pixel in the image, where the total number of rows is $m$ and columns is $n$.
        
        According to \autoref{eq:segQuality}, the leaf segmentation is based on the logical operations "$\cap$" and "$\cup$", comparing pixel to pixel of the resulting mask $A$ with the ideal mask $B$. The $Q_{seg}$ measure varies between 0 and 1, where, $Q_{seg} = 1$ indicates that the segmentation is perfect and $Q_{seg} = 0$ indicates that the segmentation is not correct.
        
        \subsubsection{Leaf Segmentation}
        
            To ensure quality in the feature extraction, especially in the calculation of the severity, which depends directly on the area of the leaf, it is necessary that the segmentation process works correctly. So, for a good result in the segmentation, the following procedures are adopted \cite{prasetyo2017mango}:
            
            \begin{enumerate}
                \item Transform the input image into the HSV or YCbCr color space.
                \item Select the best component.
                \item Apply the Otsu method \cite{otsu1979threshold} to obtain the threshold for segmentation.
                \item Apply the threshold to get the binary image.
                \item Apply the morphological operations of opening and closing for elimination of noise and filling of failures in the binary image.
            \end{enumerate}
            
\begin{figure}[H]
                \centering
                \subfloat[Leaf 1.\label{fig:m1}]
                {\includegraphics[angle=90,width=0.3\textwidth]
                    {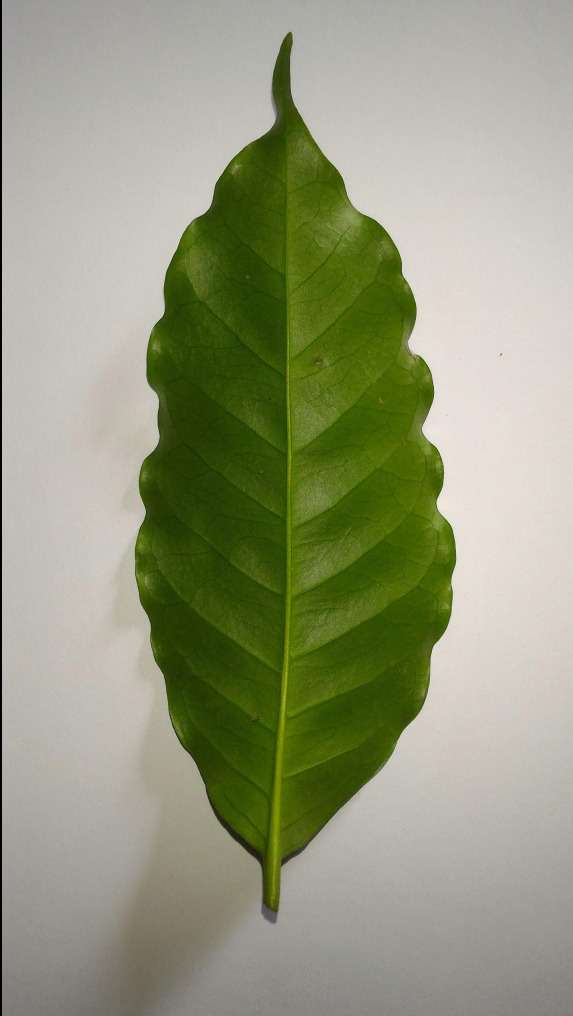}} \hspace{0.1cm}
                \subfloat[PhotoShop mask.\label{fig:m11}]
                {\includegraphics[angle=90,width=0.3\textwidth]
                    {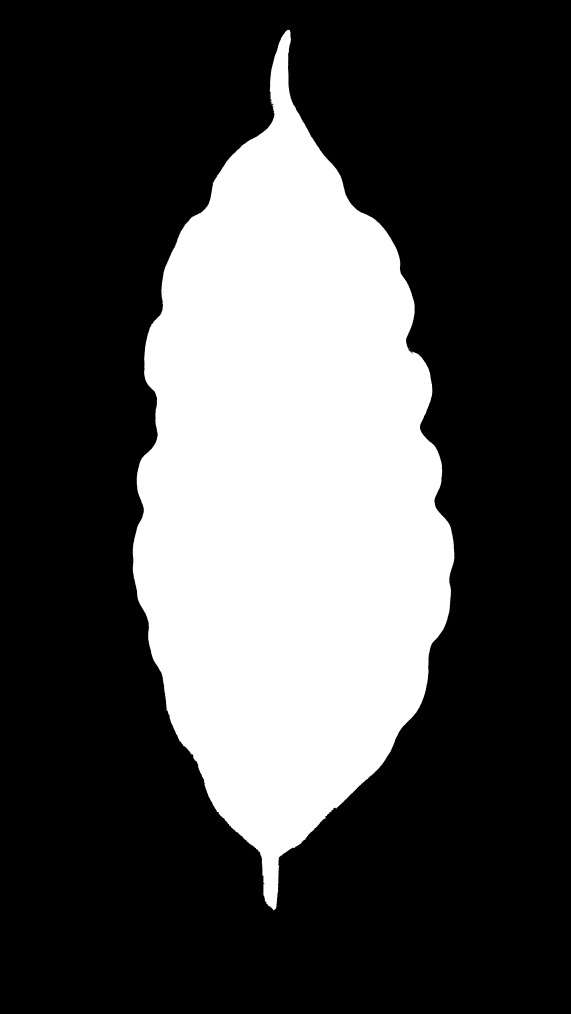}}\\
                \subfloat[Leaf 2.\label{fig:m2}]
                {\includegraphics[angle=90,width=0.3\textwidth]
                    {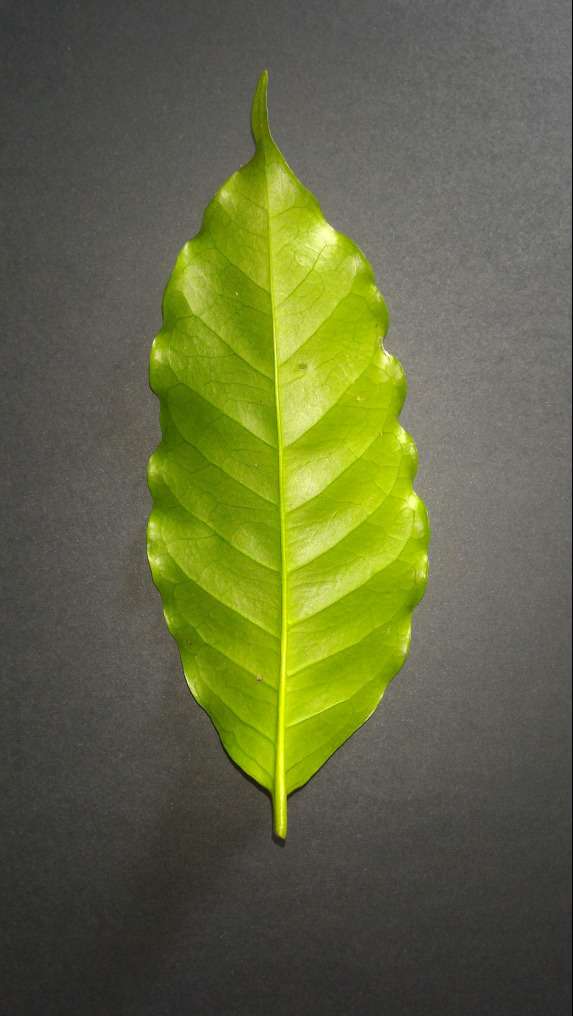}} \hspace{0.1cm}
                \subfloat[PhotoShop mask.\label{fig:m22}]
                {\includegraphics[angle=90,width=0.3\textwidth]
                    {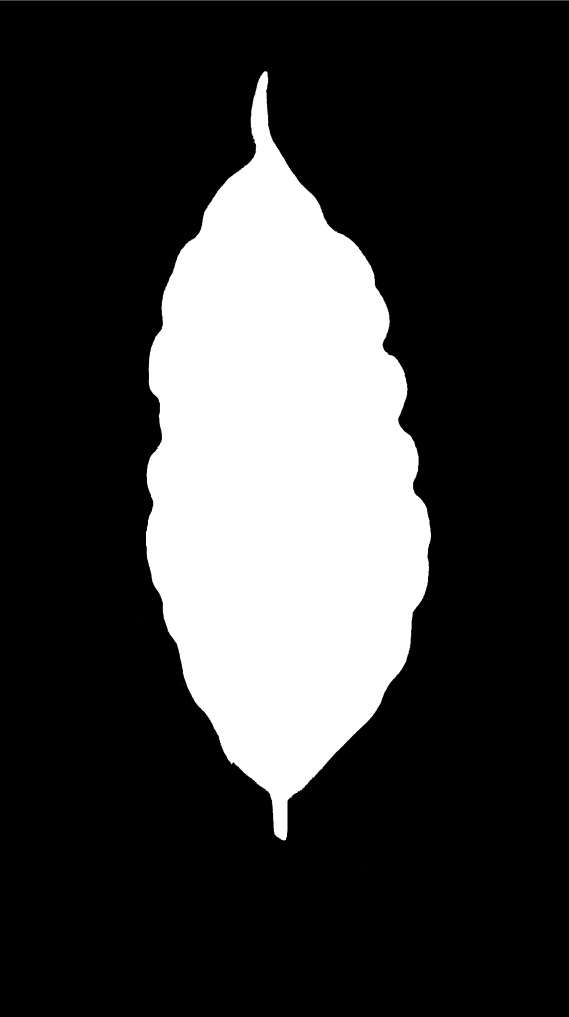}}\\
                \subfloat[Leaf 3.\label{fig:m3}]
                {\includegraphics[angle=90,width=0.3\textwidth]
                    {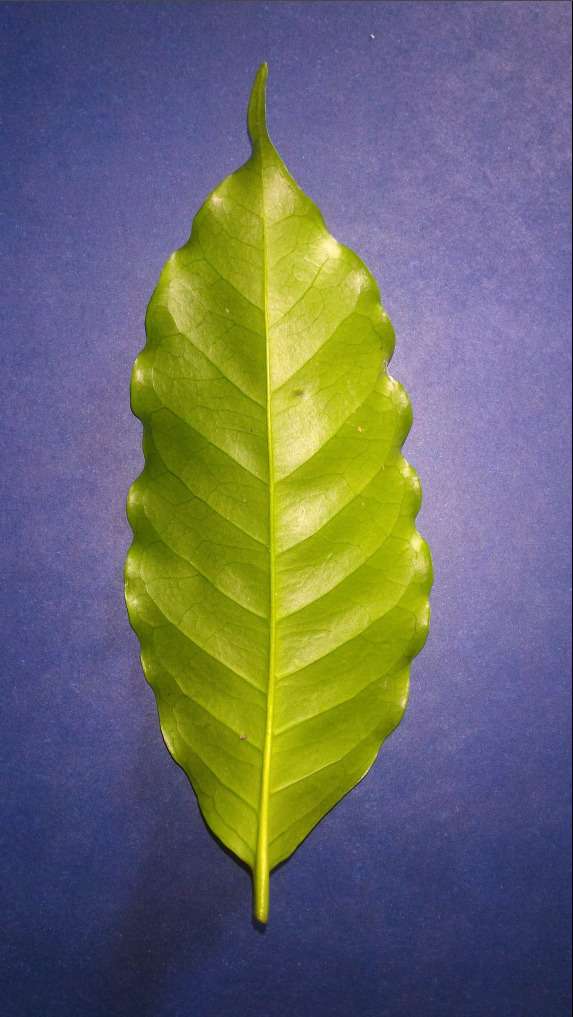}} \hspace{0.1cm}
                \subfloat[PhotoShop mask.\label{fig:m33}]
                {\includegraphics[angle=90,width=0.3\textwidth]
                    {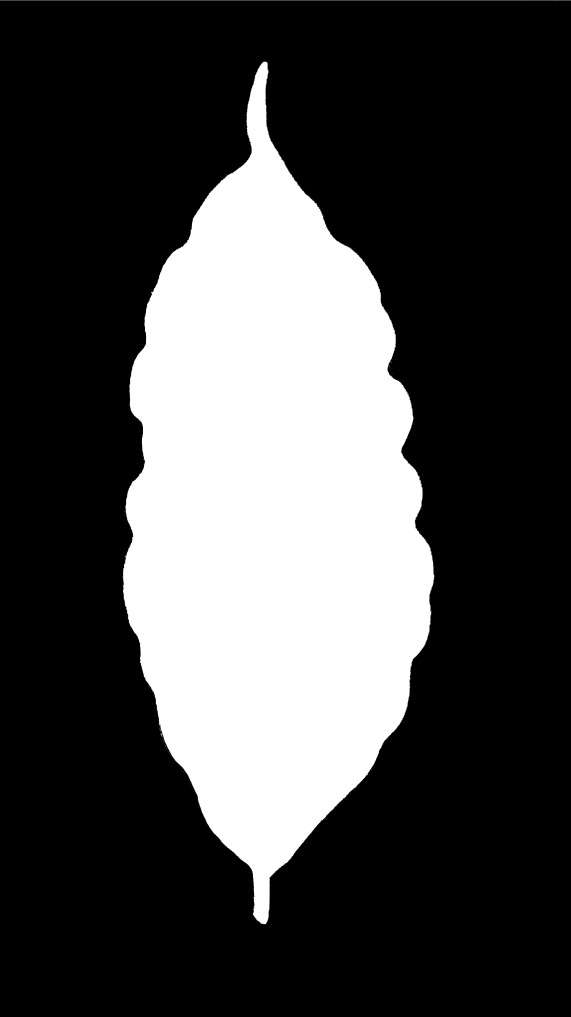}}
                \caption{Masks using PhotoShop.} 
                \label{fig:photoshop}
            \end{figure}

            To determine the best component for application of the Otsu method, it is necessary to evaluate each binary mask resulting from the segmentation in each channel. Segmentation quality is evaluated based on the metric described by \autoref{eq:segQuality}. For purpose of comparison, 20 leaf images in the three different backgrounds (white, black and blue) were chosen in order to find out the best background to be used in the segmentation process. The masks manually segmented were obtained using Photoshop as shown in \autoref{fig:photoshop}.

            %\begin{figure}[H]
            %    \centering
            %    \includegraphics[scale=0.3]{figs/18.png}
            %    \caption{\label{fig:photoshop} Masks obtained \textit{PhotoShop}.}
            %\end{figure}

            The results for segmentation accuracy are listed in \autoref{tab:QsegAval}.
                
            \begin{table}[h!]
                \centering
                \resizebox{0.5\textwidth}{!}{% <------ Don't forget this %
                \begin{tabular}{|c|c|c|c|c|c|c|} \hline
                    Background & \multicolumn{6}{c|}{Qseg} \\ \hline
                     & CbS & Cb & Cr & H & S & V \\ \hline
                    White & 0,978 & 0,961 & 0,680 & 0,746 & \textbf{0,990} & 0,845 \\ \hline
                    Blue & 0,828 & \textbf{0,970} & 0,729 & 0,770 & 0,755 & 0,018 \\ \hline
                    Black & \textbf{0,983} & 0,972 & 0,222 & 0,265 & 0,982 & 0,722 \\ \hline
                    \end{tabular}% <------ Don't forget this %
                }
                \caption{Accuracy of segmentation for different backgrounds and color components.}\label{tab:QsegAval}
            \end{table}

\begin{figure}[H]
                \centering
                \subfloat[Input image.\label{fig:i1}]
                {\includegraphics[width=0.16\textwidth]
                    {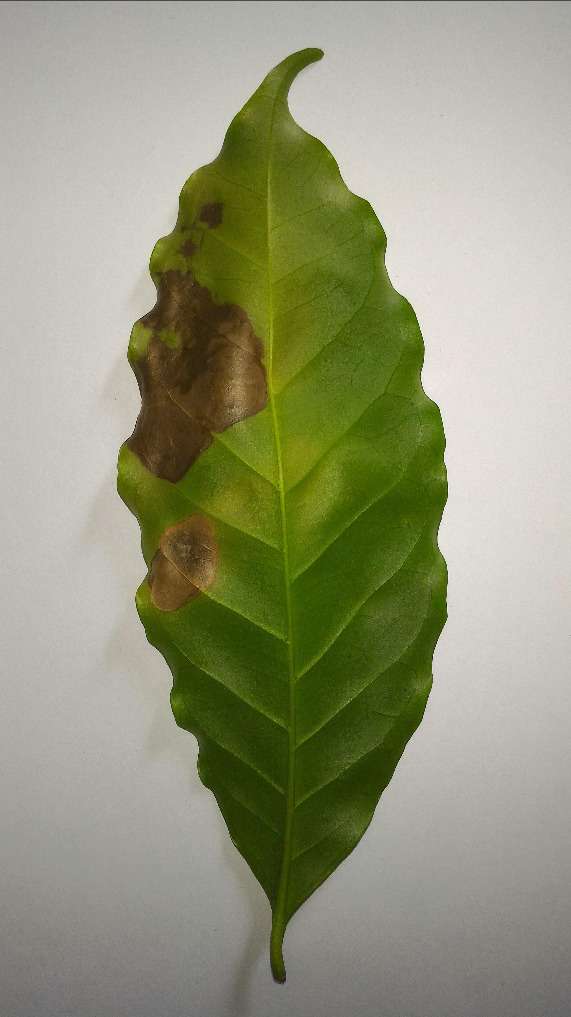}} \hspace{0.1cm}
                \subfloat[Component S.\label{fig:i2}]
                {\includegraphics[width=0.16\textwidth]
                    {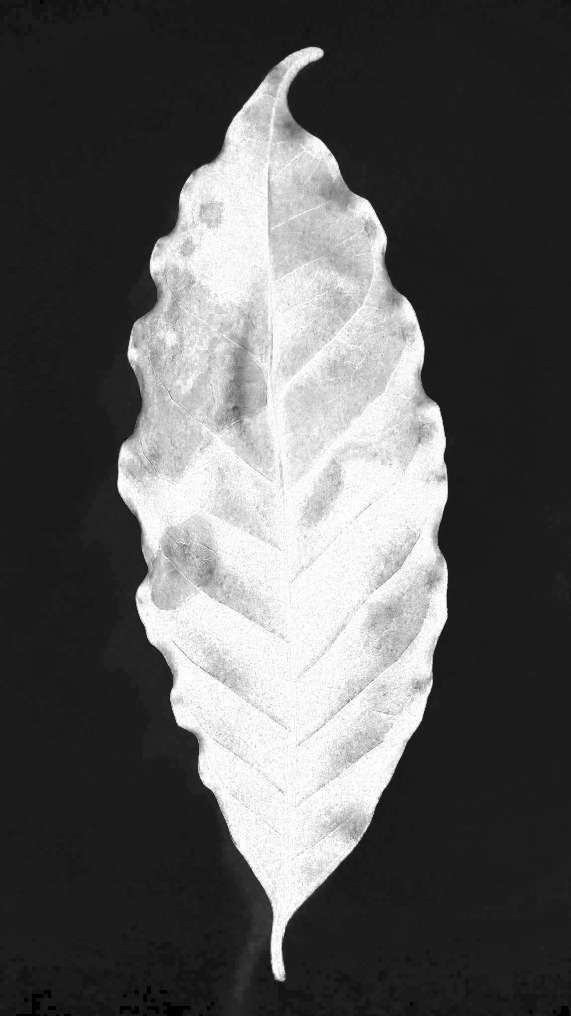}} \hspace{0.1cm}
                \subfloat[Binary mask.\label{fig:i3}]
                {\includegraphics[width=0.16\textwidth]
                    {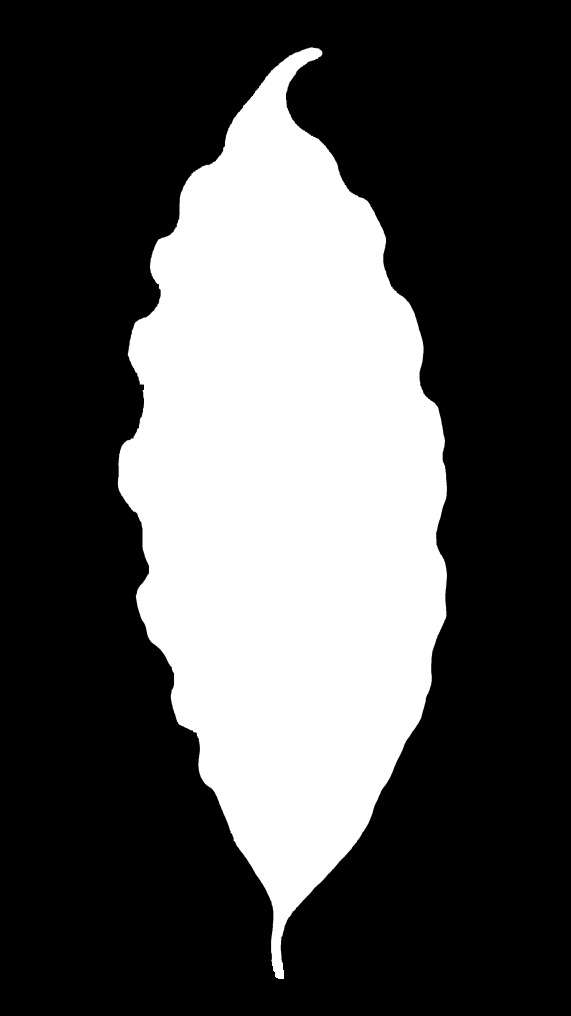}} \hspace{0.1cm}
                \subfloat[Result.\label{fig:i4}]
                {\includegraphics[width=0.16\textwidth]
                    {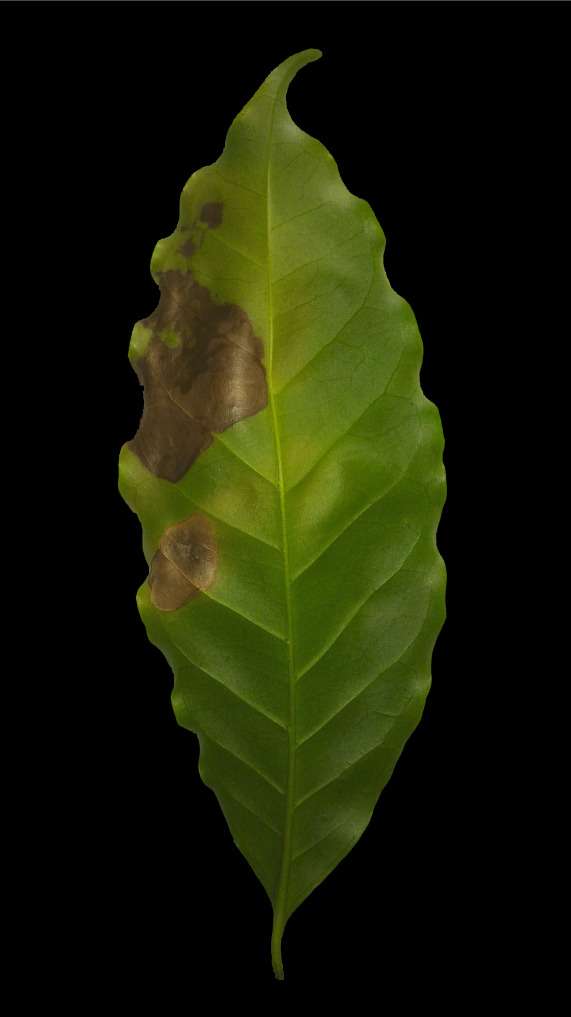}} \hspace{0.1cm}
                \caption{Images of the leaf segmentation process.} 
                \label{fig:RgbSBRgb}
            \end{figure}

            The segmentation quality depends on the background used and the color component. For example, by using the blue background, it is not recommended to use the component V but rather the component Cb. The CbS component in \autoref{tab:QsegAval} is a mixture of the Cb and S component aiming at a better segmentation. In this case, for the black background this approach was more effective. The images of the leaf segmentation process are shown in \autoref{fig:RgbSBRgb}.
            
            %\begin{figure}[H]
            %    \centering
            %    \includegraphics[scale=0.3]{figs/19.png}
            %    \caption{Images of the leaf sementation process. \label{fig:RgbSBRgb}}
            %\end{figure}
            
\begin{figure}[H]
                \centering
                \includegraphics[scale=1.4]{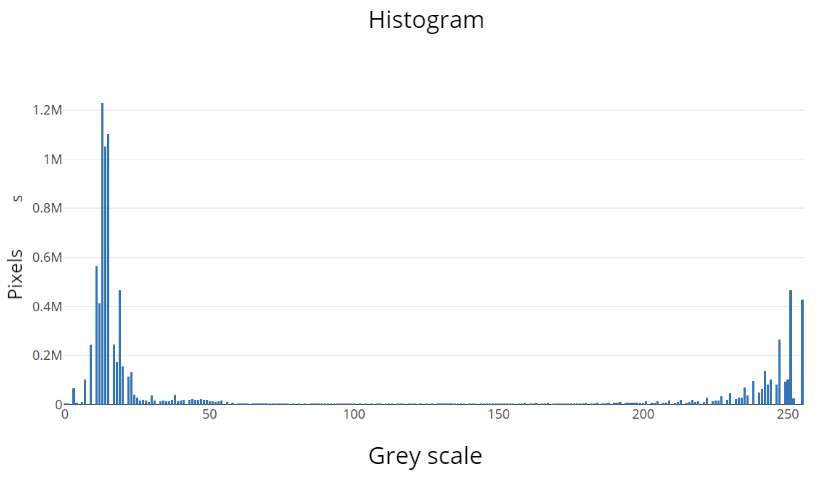}
                \caption{\label{fig:histogramaS} Histogram for the S component in HSV color space.}
            \end{figure}

             The high accuracy of the images segmentation using white background in the S component of the HSV color space is evident by analysing the histogram of this component as shown in \autoref{fig:histogramaS}. This shows that the use of the Otsu method works pretty well, since the histogram is bimodal. Therefore, determining a threshold for the separation between leaf and background was in this case quite easy.

        \subsubsection{Segmentation of injured coffee leaves}
            
\begin{figure}[H]
                \centering
                \includegraphics[scale=1.5]{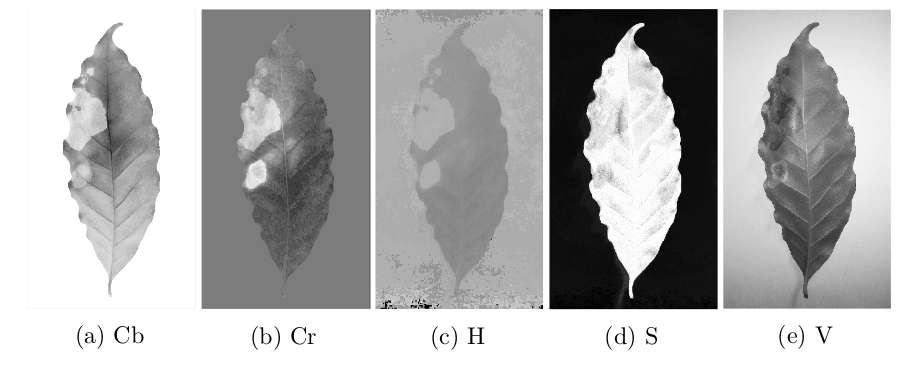}
                \caption{\label{fig:CbCrHSV} Components of the color space YCbCr and HSV.}
            \end{figure}

            After the process of segmentation of the leaf it is necessary to segment of injured coffee leaves. This step is also very important because it directly influence the quality of the attributes that are extracted from the damaged area and in the calculation of severity. The segmentation process of injured coffee leaves can be listed as follows:
                
            \begin{enumerate}
                \item Transform the input image of the leaf already segmented into the HSV and YCgCr color space, where the components are shown in \autoref{fig:CbCrHSV}.
                
                \item Apply iterative threshold segmentation algorithm in the YCgCr color space \cite{de2003face}, or apply segmentation using the \textit{k-means} algorithm in component Cr.
    
            \end{enumerate}
            
           For the segmentation of the foliar damages two algorithms are compared: the \textit{k-means} algorithm applied to the Cr component of the image converted to the YCbCr color space; or the iterative threshold segmentation method applied in the converted image to the YCgCr color space \cite{de2003face}. 
            For the iterative threshold algorithm \cite{de2003face}, the stopping criterion is reached when the difference between thresholds of 600 successive iterations is less than or equal to 0.0001. For the \textit{k-means} algorithm, the input parameter is the number of random centroids. In our case, $k=3$ was used. For values of $k$ lower than this, the injured leaf area end up not being properly separated of the rest of the image. 
            
\begin{figure}[H]
                \centering
                \subfloat[Input image.\label{fig:j1}]
                {\includegraphics[width=0.16\textwidth]
                    {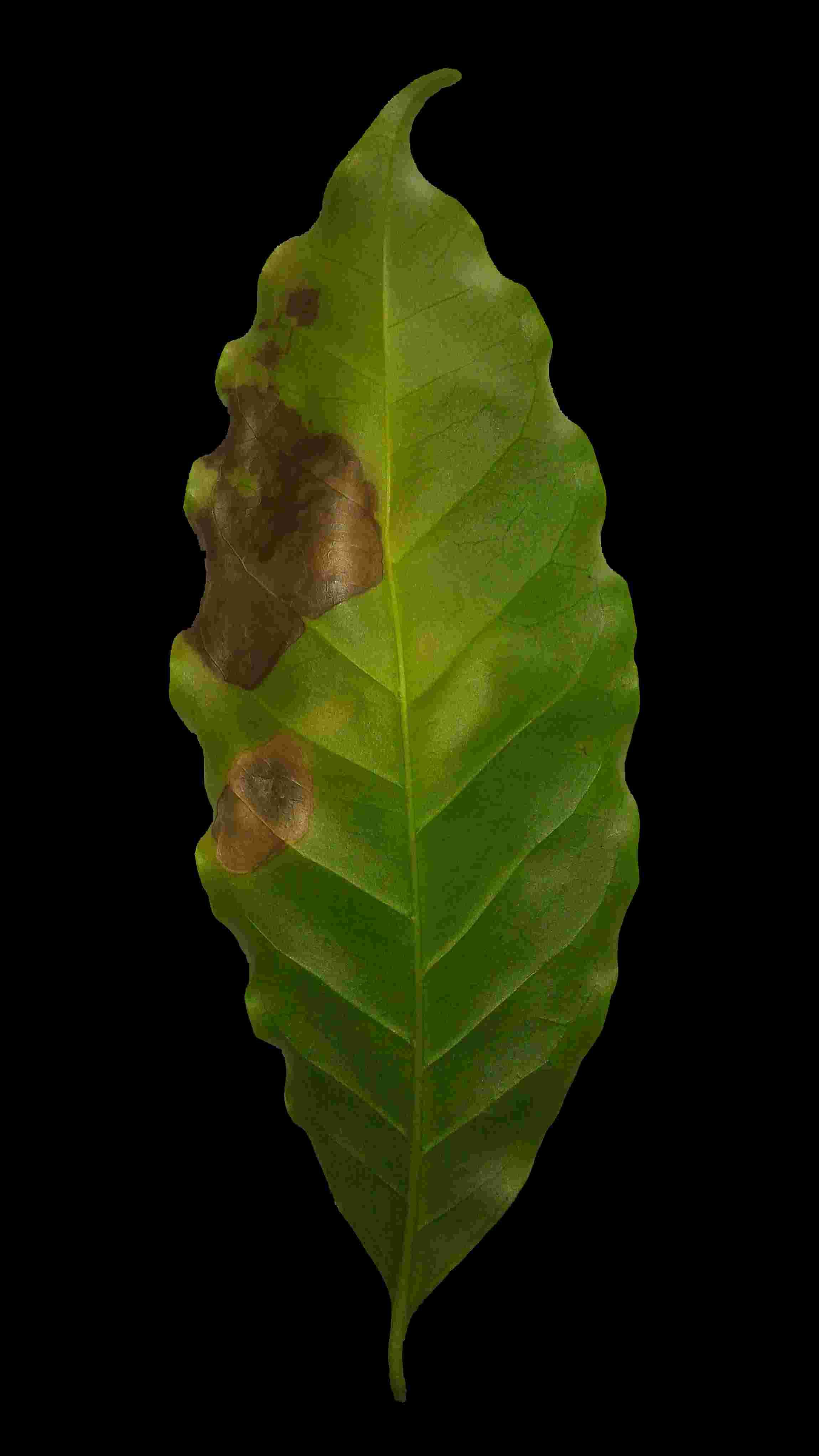}} \hspace{0.1cm}
                \subfloat[Obtained with PhotoShop.\label{fig:j2}]
                {\includegraphics[width=0.16\textwidth]
                    {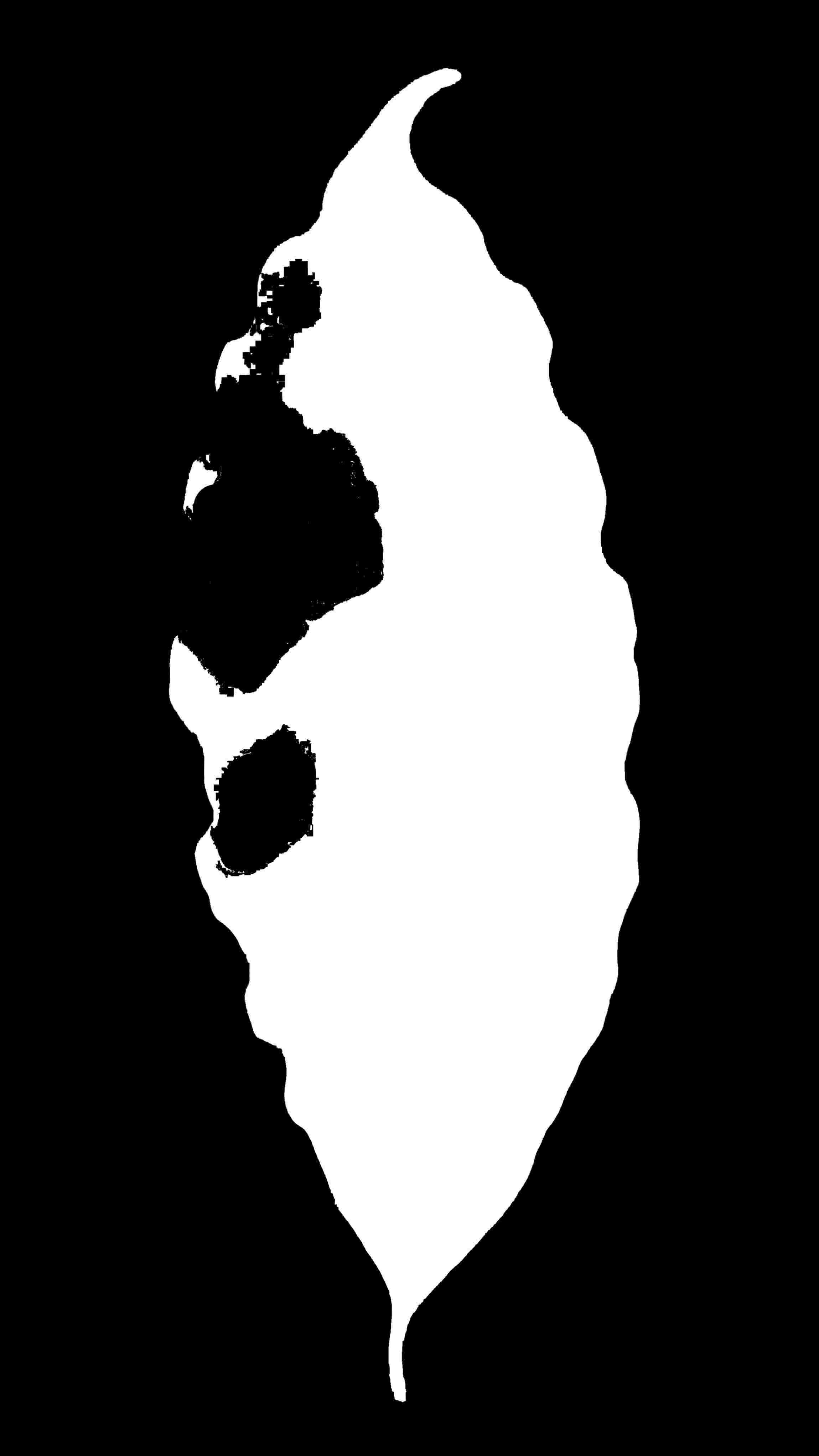}} \hspace{0.1cm}
                \subfloat[Obtained with k-means.\label{fig:j3}]
                {\includegraphics[width=0.16\textwidth]
                    {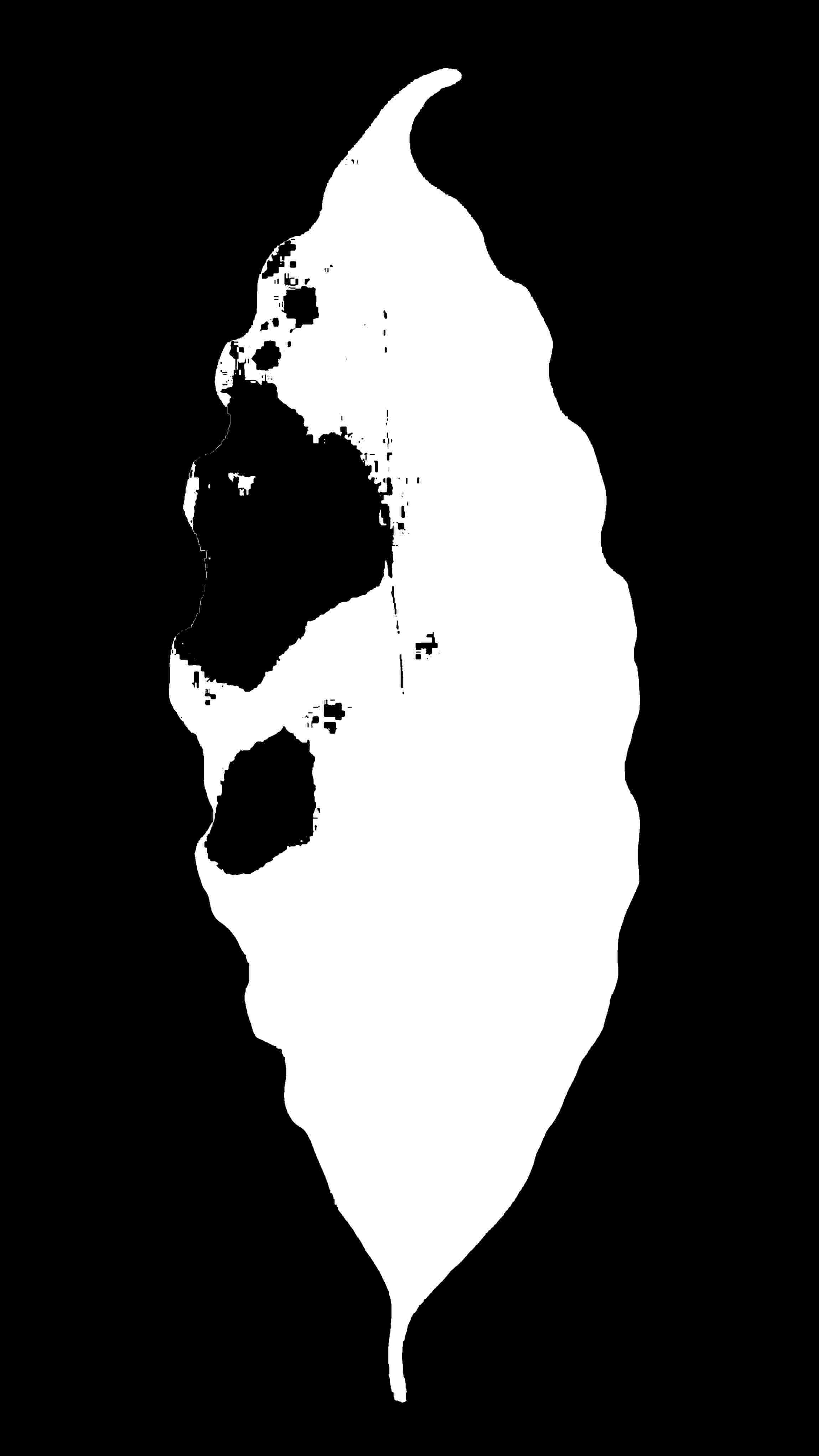}} \hspace{0.1cm}
                \subfloat[Obtained with YCgCr.\label{fig:j4}]
                {\includegraphics[width=0.16\textwidth]
                    {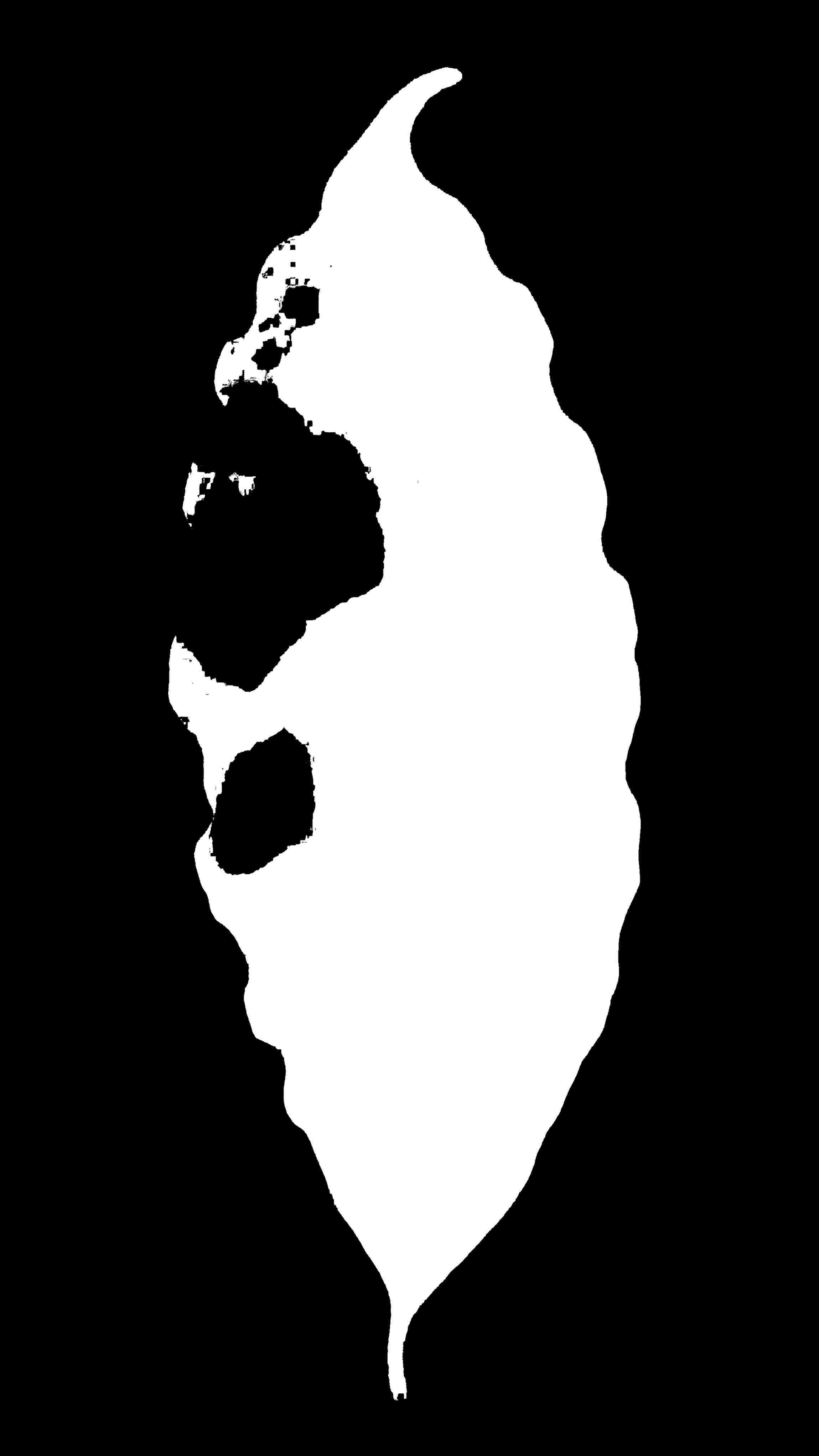}} \hspace{0.1cm}
                \caption{Masks generated from the segmented leafs.} 
                \label{fig:maskLesoes}
            \end{figure}

            In \autoref{fig:maskLesoes} is shown the target mask manually obtained through Photoshop and the other tested.
            
            %\begin{figure}[H]
            %    \centering
            %    \includegraphics[scale=0.3]{figs/23.png}
            %    \caption{\label{fig:maskLesoes} Masks generated from the segmented leafs.}
            %\end{figure}
            
            The results of segmentation of damaged foliar area for each method are presented in \autoref{tab:QsegAval2}.
                
            \begin{table}[H]
                \centering
                \resizebox{0.6\textwidth}{!}{% <------ Don't forget this %
                \begin{tabular}{|c|c|c|c|c|c|c|} \hline
                     & \multicolumn{2}{c|}{White background} & \multicolumn{2}{c|}{Blue background} & \multicolumn{2}{c|}{Black background} \\ \hline
                    Evaliation & k-means & YCgCr & k-means & YCgCr & k-means & YCgCr \\ \hline
                    Qseg & 0,975 & \textbf{0,976} & \textbf{0,949} & 0,879 & 0,487 & \textbf{0,924} \\ \hline
                    Time (s) & 15,21 & 6,32 & 16,52 & 7,10 & 14,43 & 6,52 \\ \hline
                    \end{tabular}% <------ Don't forget this %
                }
                \caption{Comparison between k-means algorithm and YCgCr for different backgrounds.}\label{tab:QsegAval2}
            
            \end{table}
            
\begin{figure}[H]
                \centering
                \subfloat[Input image.\label{fig:h1}]
                {\includegraphics[width=0.16\textwidth]
                    {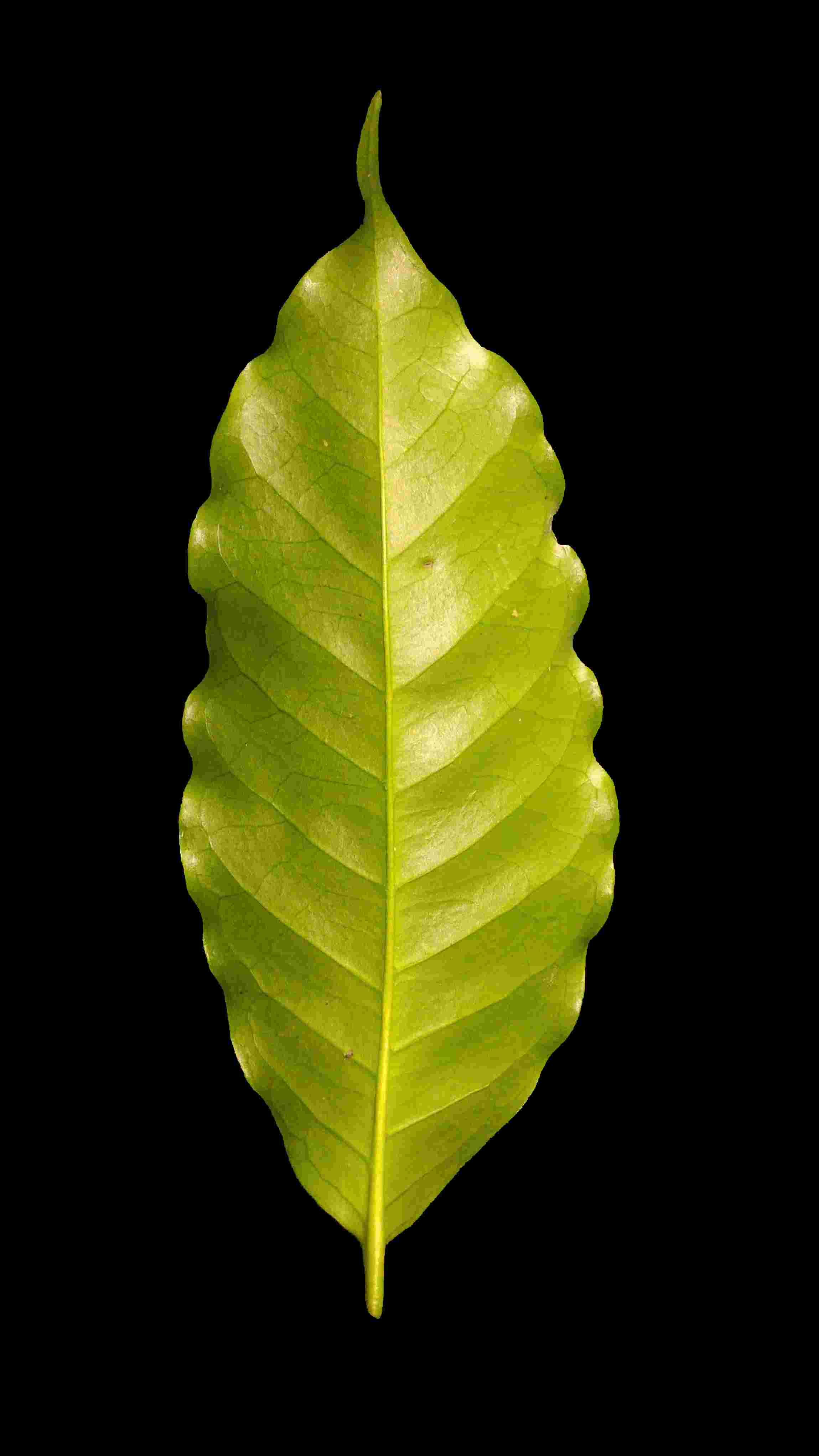}} \hspace{0.1cm}
                \subfloat[Obtained with YCgCr.\label{fig:h2}]
                {\includegraphics[width=0.16\textwidth]
                    {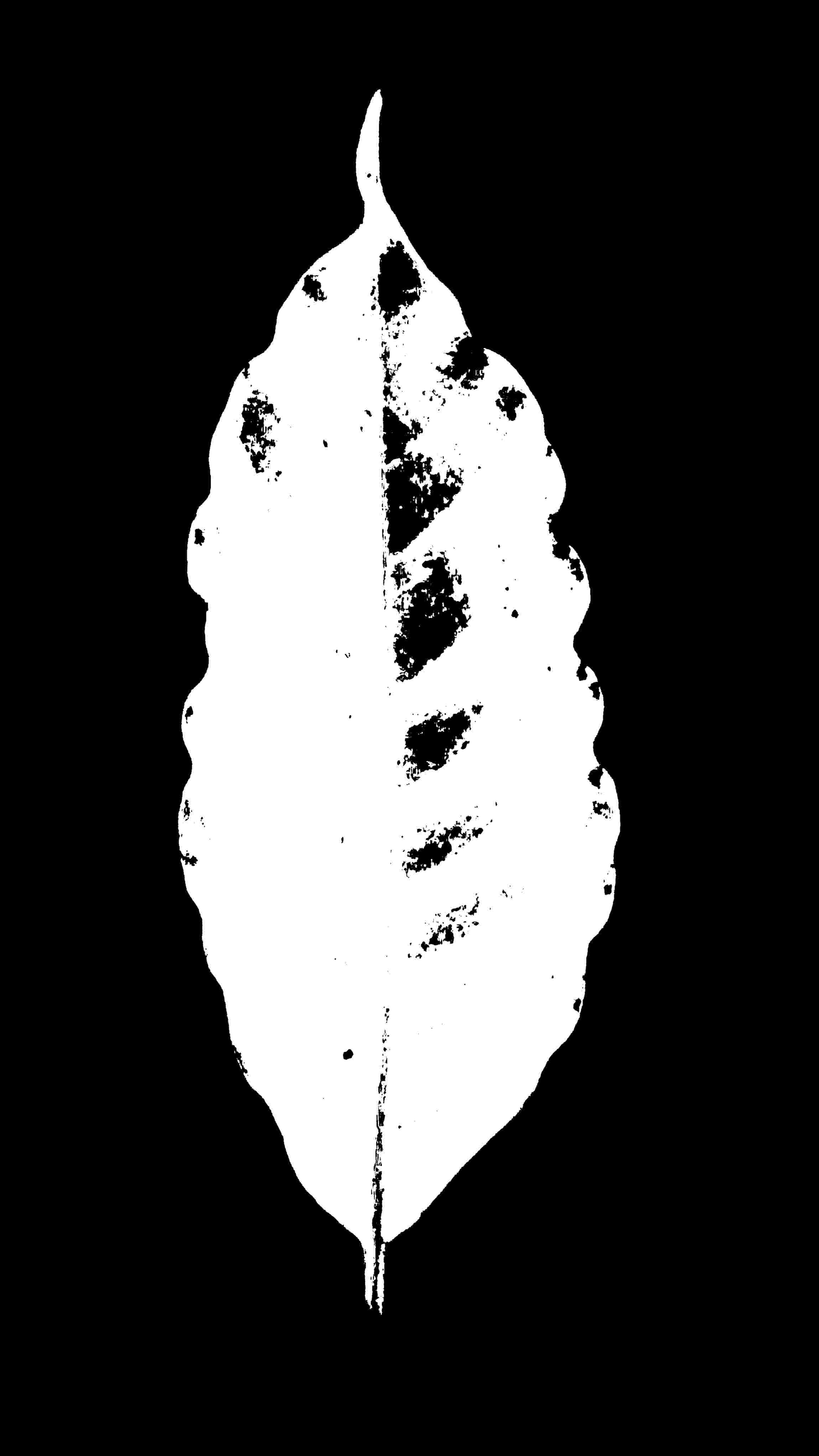}} \hspace{0.1cm}
                \subfloat[Result.\label{fig:h3}]
                {\includegraphics[width=0.16\textwidth]
                    {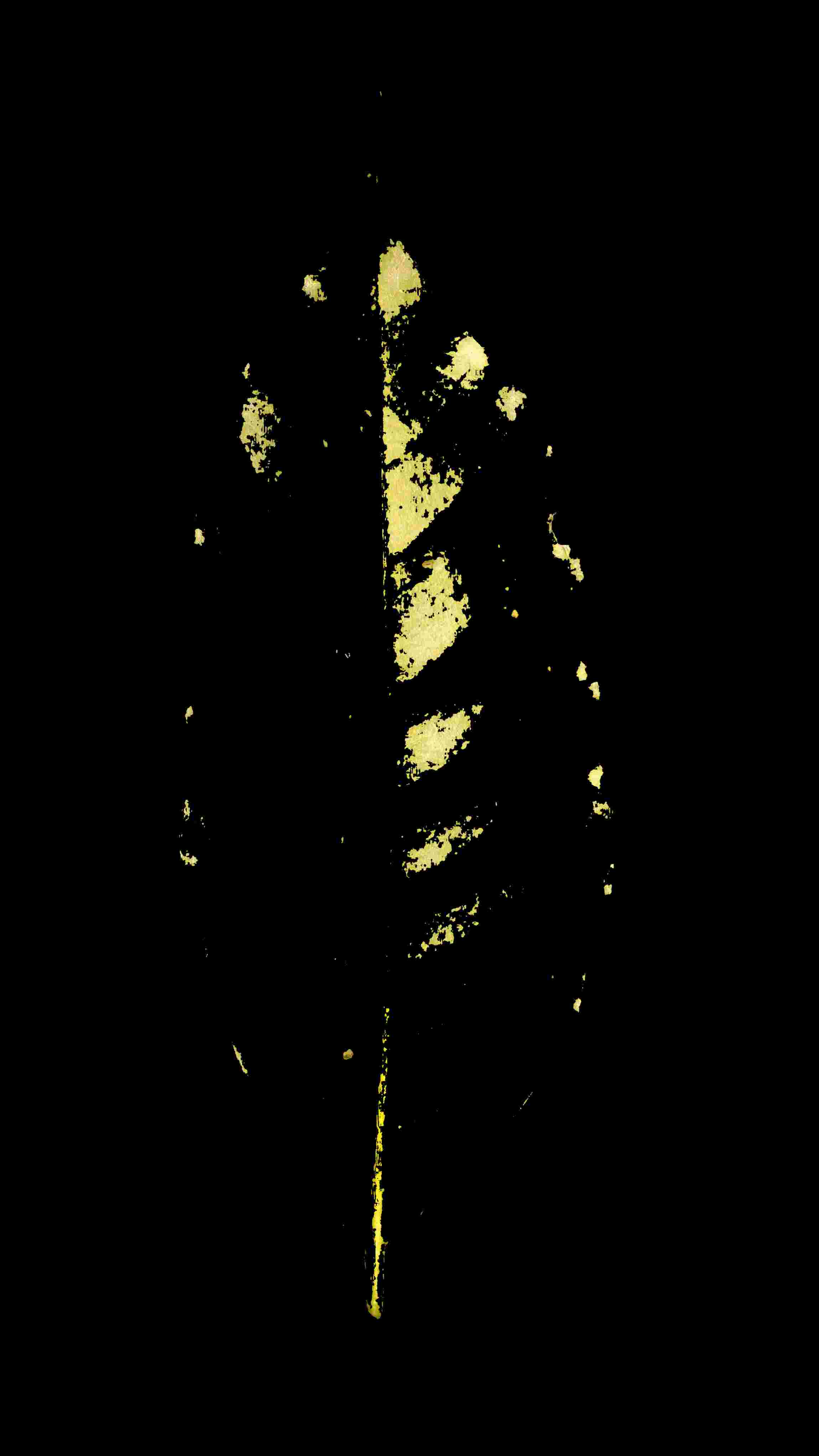}} \hspace{0.1cm}
                \caption{A case showing the occurrence of error in the process of injured leaf segmentation.} 
                \label{fig:falha}
            \end{figure}

\begin{figure}[H]
                \centering
                \subfloat[Input image.\label{fig:g1}]
                {\includegraphics[width=0.16\textwidth]
                    {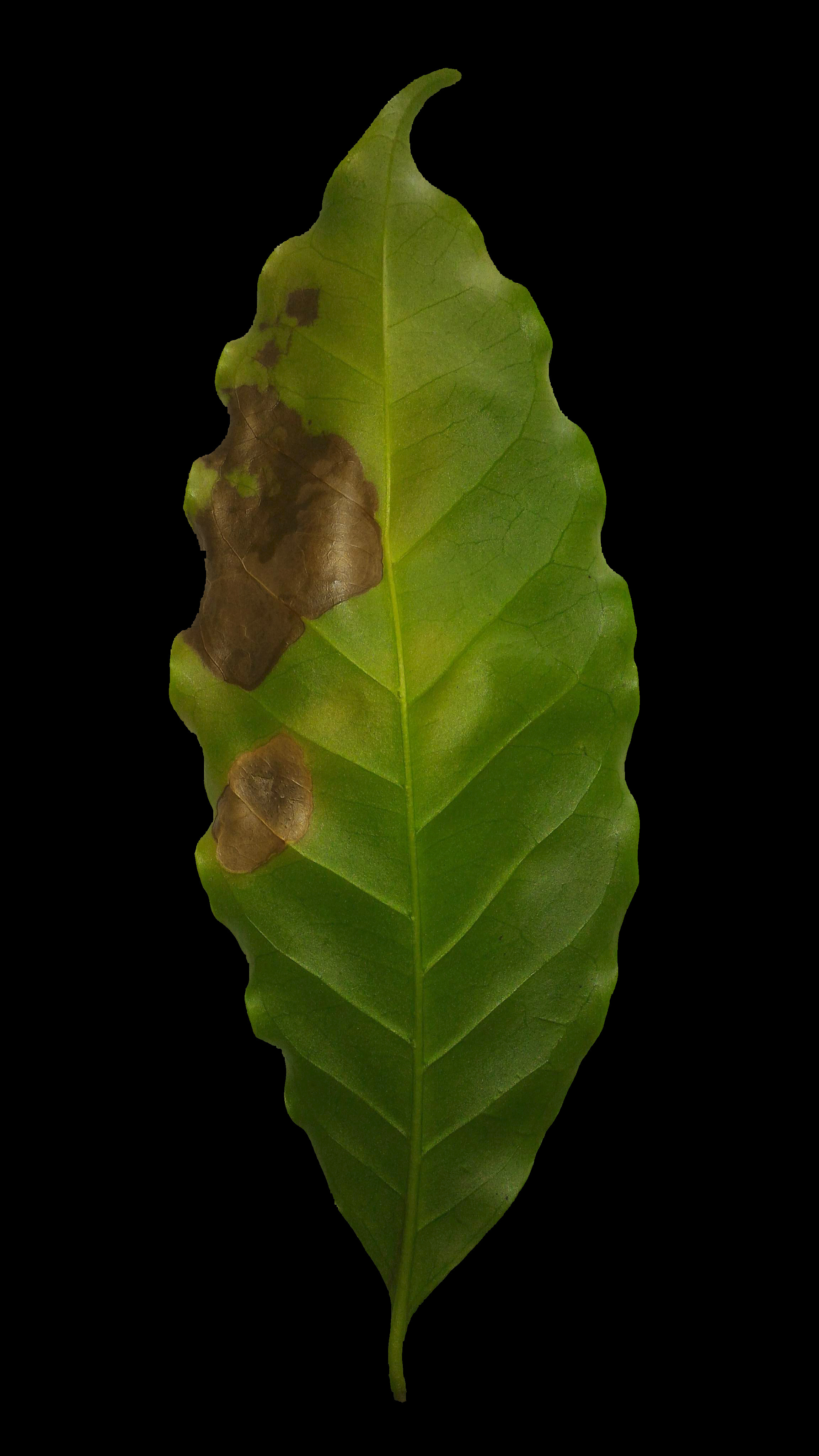}} \hspace{0.1cm}
                \subfloat[Obtained with YCgCr.\label{fig:g2}]
                {\includegraphics[width=0.16\textwidth]
                    {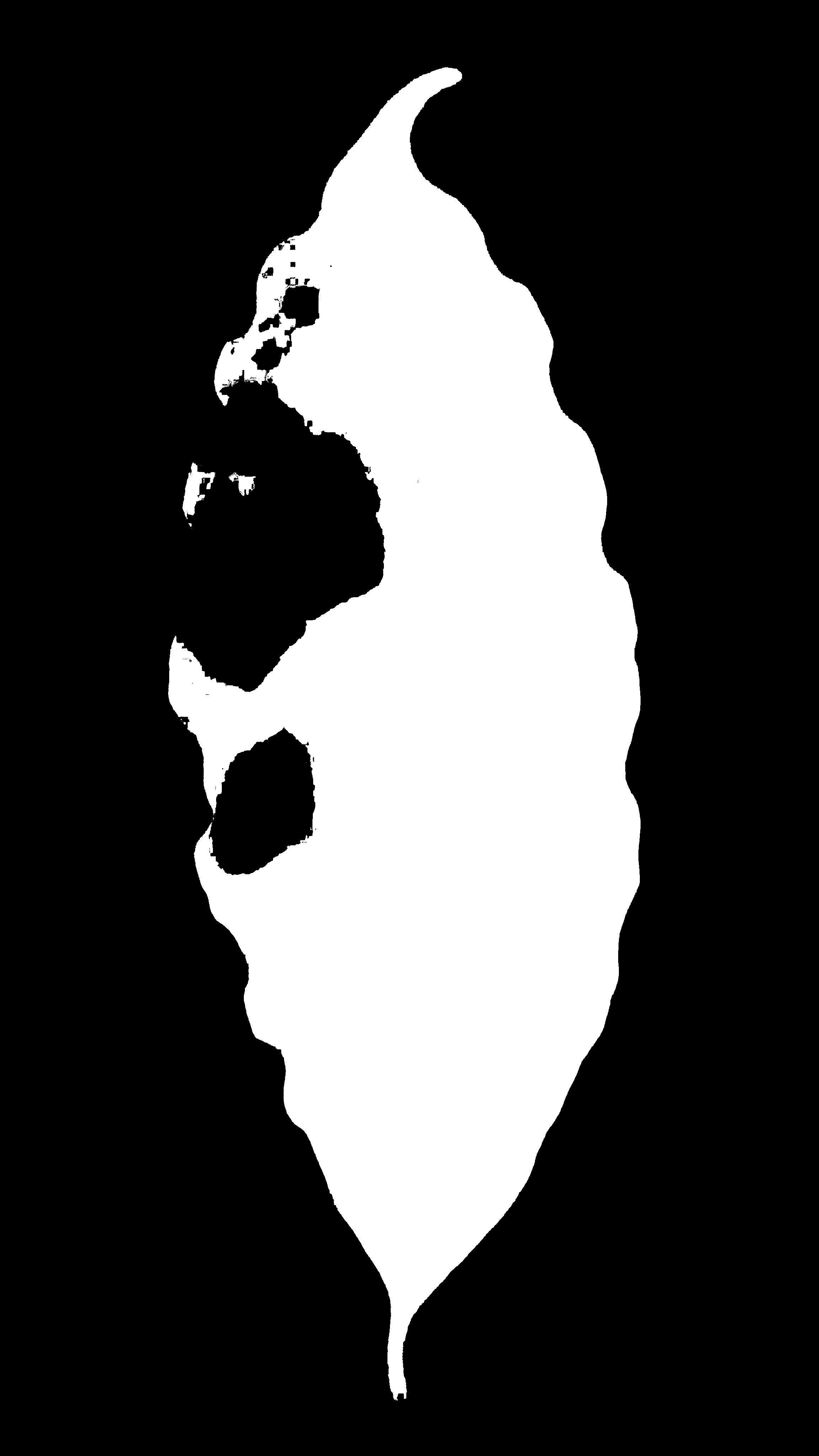}} \hspace{0.1cm}
                \subfloat[Result.\label{fig:g3}]
                {\includegraphics[width=0.16\textwidth]
                    {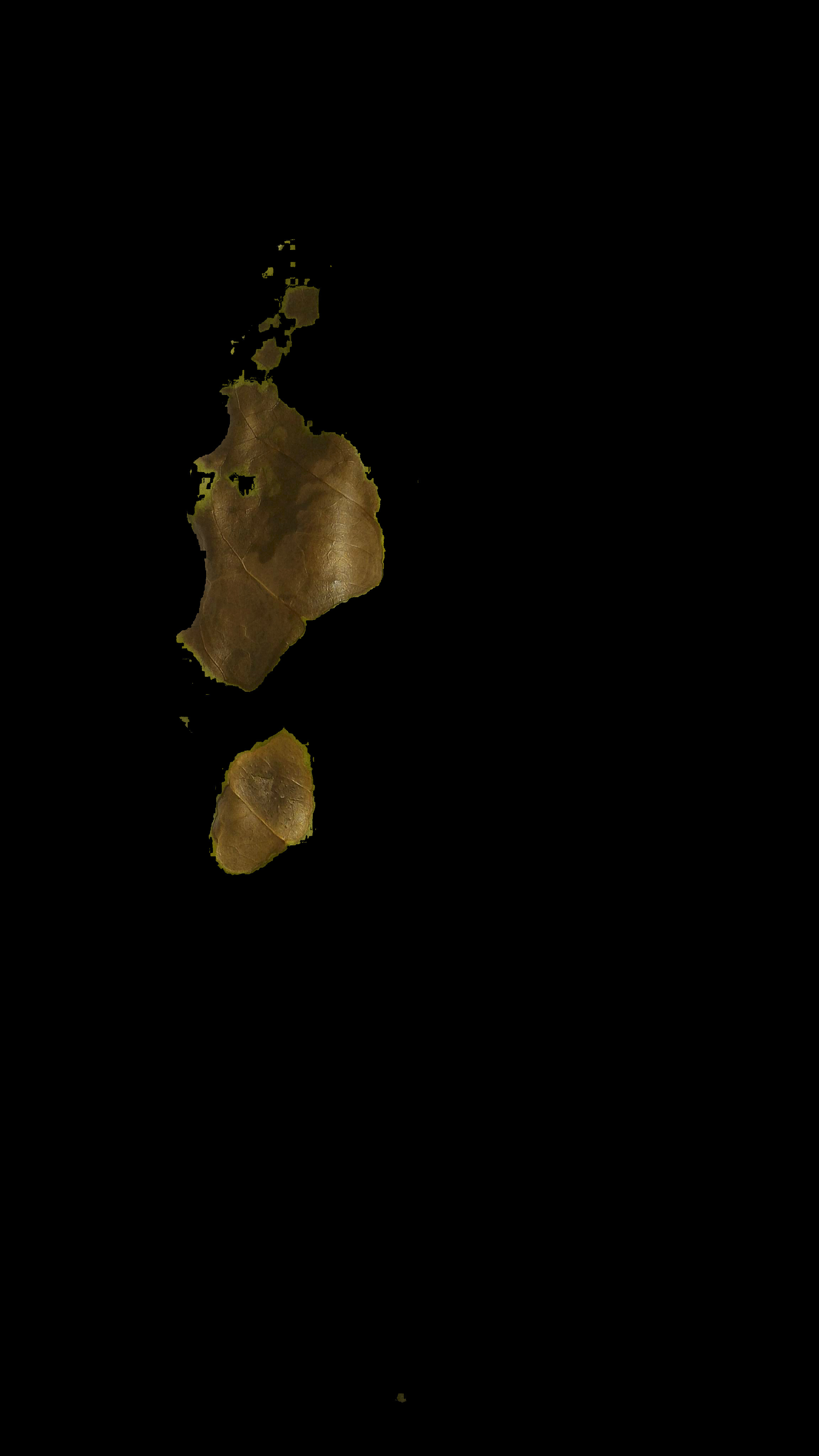}} \hspace{0.1cm}
                \caption{A sussesfull case in the process of injured leaf segmentation.} 
                \label{fig:lesoesSegFinal}
            \end{figure}

            It can be observed that the method using iterative threshold in YCgCr color space is more accurate in the segmentation and also faster than the \textit{k-means}. The white background also shows better results for injured leaf separation. This was mainly because the camera, at the time of capture, was in the automatic color adjustment mode. Therefore, the different backgrounds generate different color tones in the coffee leaf.  In addition, the incidence of flash creates reflective focus on the leaf. \autoref{fig:falha} shows one of these segmentation errors caused by these anomalies, where the segmentation method encounters damaged foliar areas that did not exist. On the other hand, in \autoref{fig:lesoesSegFinal}, is illustrated a case where the segmentation occurred perfectly.
            
            %\begin{figure}[H]
            %    \centering
            %    \includegraphics[scale=0.3]{figs/24.png}
            %    \caption{\label{fig:falha} A case with error in the process of lesions segmentation.}
            %\end{figure}

            %\begin{figure}[H]
            %    \centering
            %    \includegraphics[scale=0.3]{figs/25.png}
            %    \caption{A sussesfull case in the process of lesion segmentation. %\label{fig:lesoesSegFinal}}
            %\end{figure}
            
            By mean of these investigations, one can conclude that the best background would be white, since it presents better results to the leaf segmentation as well as to the segmentation of the leaf injured foliar area.

    \subsection{Classification results}
    
       In this work, artificial neural network trained with Backpropagation algorithms and extreme learning machine was used. In both cases the best architecture were evaluated according to the most used metrics: sensitivity and precision \cite{powers2011evaluation}.
       
       \subsubsection{Metrics used}
        
        \begin{itemize}
            \item \textbf{Recall:} Denotes the proportion of true positives over the true positives and false negatives described as:
            
            \begin{equation}
        		\begin{aligned}
        		    Recall = \frac{tp}{tp+fn}
        		\end{aligned}
        		\label{eq:reacall}
    	    \end{equation}

            \noindent where $tp$ is the number of true positives and $fn$ is the amount of false negatives.
            
            \item \textbf{Precision:} Denotes the proportion of true positives over all positive inferences and described as:

            \begin{equation}
        		\begin{aligned}
        		    Precision = \frac{tp}{tp+fp}
        		\end{aligned}
        		\label{eq:precisao}
    	    \end{equation}

            \noindent where $tp$ is the number of true positives and $fp$ of false positives.
            
            \item \textbf{Accuracy:} Denotes the proportion of true results (both true positives and true negatives) over the total number of cases examined described as:
            
            \begin{equation}
        		\begin{aligned}
        		    Accuracy = \frac{tp+tn}{total}
        		\end{aligned}
        		\label{eq:accuracy}
    	    \end{equation}
    	    
    	    \noindent where $tp$ is the number of true positives, $tn$ is the amount of true negatives and $total$ is the number of cases examined.
            
        \end{itemize}

    \subsubsection{Classification results using Backpropagation}
    
        In the neural network training the following setup and parameters were used:  Stochastic gradient descent (sgd) was used to minimize the error \cite{bottou2010large}. Activation function: ReLu was used as it is the most frequently used due to rapid learning \cite{lecun2015deep}.
        The learning rate tested was 0.001, 0.01; 0.1; 0.2 and 0.3. The stop criterion: the training is interrupted when the difference between the error of two successive iterations is less than 0.0001. Decay: Determines the decay learning rate along the training period. We use a constant decay  as well as exponential decay over time. The number of neurons used in the hidden layer started with 10 neurons.
        
        For each configuration, the training was performed 100 times. To each new training, testing and training sets were random chosen, with $70\%$ of data for training and $30\%$ for testing. \autoref{tab:testesANN}  shows the configurations that obtained the best results.
        
        \begin{table}[H]
            \centering
            \resizebox{0.8\textwidth}{!}{% <------ Don't forget this %
                \begin{tabular}{|l|c|c|c|l|c|c|}
                    \hline
                     & \multicolumn{3}{c|}{Parameters} &  & \multicolumn{1}{l|}{Results} & \multicolumn{1}{l|}{} \\ \hline
                    Model & \begin{tabular}[c]{@{}c@{}}Neurons in\\ the hidden \\ layer\end{tabular} & \begin{tabular}[c]{@{}c@{}}Learning \\ rate\end{tabular} & \begin{tabular}[c]{@{}c@{}}Update\\ of the \\ learning rate\end{tabular} & \multicolumn{1}{c|}{Accuracy (\%)} & Precision (\%) & Recall (\%) \\ \hline
                    1 & 20 & 0.1 & Constant & 95.527 $\pm$ 0.945 & 95.906 $\pm$ 0.949 & 93.428 $\pm$ 0.924 \\ \hline
                    2 & 30 & 0.1 & Constant & 95.824 $\pm$ 0.948 & 95.959 $\pm$ 0.949 & 93.142 $\pm$ 0.922 \\ \hline
                    3 & 40 & 0.1 & Constant & 96.013 $\pm$ 0.950 & 96.151 $\pm$ 0.951 & 93.233 $\pm$ 0.922 \\ \hline
                    4 & 50 & 0.1 & Constant & 95.726 $\pm$ 0.947 & 95.970 $\pm$ 0.950 & 92.688 $\pm$ 0.918 \\ \hline
                    5 & 100 & 0.1 & Constant & 95.815 $\pm$ 0.948 & 96.191 $\pm$ 0.952 & 93.181 $\pm$ 0.922 \\ \hline
                    6 & 200 & 0.1 & Constant & 95.958 $\pm$ 0.950 & 96.053 $\pm$0.950 & 92.766 $\pm$ 0.918 \\ \hline
                    7 & 10 & 0.2 & Constant & 95.175 $\pm$ 0.942 & 95.277 $\pm$ 0.943 & 92.363 $\pm$ 0.914 \\ \hline
                    8 & 20 & 0.2 & Constant & 95.041 $\pm$ 0.940 & 95.501 $\pm$ 0.945 & 92.857 $\pm$ 0.920 \\ \hline
                    9 & 30 & 0.2 & Constant & 95.027 $\pm$ 0.940 & 95.526 $\pm$ 0.945 & 93.103 $\pm$ 0.922 \\ \hline
                    10 & 40 & 0.2 & Constant & 95.212 $\pm$ 0.942 & 95.098 $\pm$ 0.941 & 92.311 $\pm$ 0.914 \\ \hline
                    11 & 50 & 0.2 & Constant & 95.307 $\pm$ 0.943 & 94.728 $\pm$ 0.937 & 93.000 $\pm$ 0.920 \\ \hline
                    12 & 100 & 0.2 & Constant & 95.136 $\pm$ 0.941 & 94.882 $\pm$ 0.939 & 93.155 $\pm$ 0.921 \\ \hline
                    13 & 200 & 0.2 & Constant & 95.220 $\pm$ 0.942 & 95.324 $\pm$ 0.943 & 93.025 $\pm$ 0.920 \\ \hline
                    \end{tabular}% <------ Don't forget this %
            }
            \caption{Comparison of neural network models with different setup and parameters.}\label{tab:testesANN}
        \end{table}
        
        One notices that the best models for this training set are those that use the constant learning rate for values between 0.1 and 0.2 with the number of neurons varying from 10 to 100.
        
    \subsubsection{Classification results with extreme learning machine}
    
        For training with ELM, the setup and parameters used were  Activation function: Linear, sigmoid and hyperbolic tangent functions. The number of neurons 10, 20, 30, 40, 50, 100 and 200 were tested in the hidden layer. The number of runs for each configuration and the database were the same as described in previous case. \autoref{tab:testesELM} shows the results of the runs for each configuration.
        
        \begin{table}[H]
            \centering
            \resizebox{0.9\textwidth}{!}{% <------ Don't forget this %
            \begin{tabular}{|c|c|c|l|c|c|} \hline & \multicolumn{2}{c|}{Parameters} & \multicolumn{3}{c|}{Results} \\ \hline
                Models & Neurons in the hidden layer & Activation func. & Accuracy (\%) & Precision (\%) & Recall (\%) \\ \hline
                1 & 10 & Linear & 95.815 $\pm$ 0.013 & 98.160 $\pm$ 0.199 & 97.350 $\pm$ 0.507 \\ \hline
                2 & 20 & Linear & 96.363 $\pm$ 0.012 & 98.099 $\pm$ 0.311 & 97.519 $\pm$ 0.312 \\ \hline
                3 & 30 & Linear & 96.351 $\pm$ 0.028 & 98.592 $\pm$ 0.090 & 97.103 $\pm$ 0.247 \\ \hline
                4 & 40 & Linear & 96.386 $\pm$ 0.024 & 98.363 $\pm$ 0.100 & 97.460 $\pm$ 0.182 \\ \hline
                5 & 50 & Linear & 96.320 $\pm$ 0.038 & 98.390 $\pm$ 0.193 & 97.337 $\pm$ 0.052 \\ \hline
                6 & 100 & Linear & 96.348 $\pm$ 0.013 & 98.590 $\pm$ 0.153 & 97.142 $\pm$ 0.455 \\ \hline
                7 & 200 & Linear & 96.357 $\pm$ 0.036 & \textbf{98.644} $\pm$ \textbf{0.226} & \textbf{97.649} $\pm$ \textbf{0.325} \\ \hline
                8 & 10 & Sigmoid & 92.523 $\pm$ 0.255 & 94.795 $\pm$ 0.634 & 87.324 $\pm$ 0.442 \\ \hline
                9 & 20 & Sigmoid & 95.504 $\pm$ 0.120 & 97.364 $\pm$ 0.322 & 94.480 $\pm$ 0.065 \\ \hline
                10 & 30 & Sigmoid & 96.958 $\pm$ 0.062 & 97.802 $\pm$ 0.475 & 96.207 $\pm$ 0.468 \\ \hline
                11 & 40 & Sigmoid & 97.258 $\pm$ 0.042 & 97.710 $\pm$ 0.027 & 97.129 $\pm$ 0.260 \\ \hline
                12 & 50 & Sigmoid & 97.298 $\pm$ 0.035 & 97.554 $\pm$ 0.298 & 97.493 $\pm$ 0.078 \\ \hline
                13 & 100 & Sigmoid & 98.128 $\pm$ 0.045 & 96.442 $\pm$ 0.277 & 96.701 $\pm$ 0.363 \\ \hline
                14 & 200 & Sigmoid & \textbf{99.095} $\pm$ \textbf{0.007} & 93.816 $\pm$ 0.427 & 92.818 $\pm$ 0.428 \\ \hline
                15 & 10 & Tang. Hiper. & 91.458 $\pm$ 0.226 & 93.619 $\pm$ 0.110 & 86.818 $\pm$ 0.377 \\ \hline
                16 & 20 & Tang. Hiper. & 94.440 $\pm$ 0.043 & 95.349 $\pm$ 0.379 & 91.909 $\pm$ 0.013 \\ \hline
                17 & 30 & Tang. Hiper. & 95.759 $\pm$ 0.006 & 96.397 $\pm$ 0.046 & 94.181 $\pm$ 0.039 \\ \hline
                18 & 40 & Tang. Hiper. & 96.472 $\pm$ 0.075 & 96.528 $\pm$ 0.062 & 95.519 $\pm$ 0.182 \\ \hline
                19 & 50 & Tang. Hiper. & 96.965 $\pm$ 0.037 & 97.075 $\pm$ 0.254 & 95.740 $\pm$ 0.792 \\ \hline
                20 & 100 & Tang. Hiper. & 97.988 $\pm$ 0.987 & 96.871 $\pm$ 0.077 & 96.272 $\pm$ 0.351 \\ \hline
                21 & 200 & Tang. Hiper. & 99.062 $\pm$ 0.003 & 93.404 $\pm$ 0.536 & 92.675 $\pm$ 0.584 \\ \hline
                \end{tabular}% <------ Don't forget this %
            }
            \caption{Comparison between ELM models.}\label{tab:testesELM}
        \end{table}
        
        The models that use the linear activation function presented, in general, better results for this base. However, we can observe that from 10 to 200 neurons in the models 1-7 the performance improvement was not significant. Another interesting point was the confirmation that the accuracy is not sufficient to evaluate the performance of the classifier, since the model 14 presents a very high accuracy but with lowest values of precision and sensitivity compared to other models. This is probably because the network becomes very specialized for the training set.      
        
        Since ELM  presents better results when compared to backpropagation, it was used as a classifier for the damaged foliar area classification and severity calculation.
        
    \subsection{Severity calculation results}

        Some works such as in\cite{mwebaze2016machine}, \cite{rastogi2015leaf} and \cite{baghel2016k} presented the calculation of severity but do not use metrics to validate the result. Therefore, we calculated the severity of the manually generated masks with the PhotoShop and compare it with the severity calculation presented in this work. The results of the severity calculation for each one of the methods are presented in \autoref{tab:resultSeve}.
        
        \begin{table}[H]
            \centering
            \resizebox{0.3\textwidth}{!}{% <------ Don't forget this %
            \begin{tabular}{|c|c|c|c|} \hline & \multicolumn{3}{c|}{Severity (\%)} \\ \hline
                Image & k-means & YCgCr & PhotoShop \\ \hline
                1 & 15,05 & 14,29 & 15,05 \\ \hline
                2 & 2,05 & 1,98 & 2,17 \\ \hline
                3 & 2,98 & 3,62 & 2,81 \\ \hline
                4 & 6,25 & 6,97 & 6,00 \\ \hline
                5 & 10,45 & 3,93 & 11,15 \\ \hline
                6 & 7,87 & 6,73 & 9,95 \\ \hline
                7 & 0,89 & 0,91 & 0,62 \\ \hline
                8 & 6,49 & 6,62 & 6,18 \\ \hline
                9 & 2,80 & 2,72 & 2,69 \\ \hline
                10 & 1,64 & 1,67 & 1,42 \\ \hline
                \end{tabular}% <------ Don't forget this %
            }
            \caption{\label{tab:resultSeve} Results for severity estimation.}
        \end{table}
        
        The difference between the severity percentages is acceptable because the assessment of severity is usually done between ranges of values \cite{rastogi2015leaf}, as presented in \autoref{tab:riscoSeve}. In addition, the segmentation done manually is an approximation of what the author considers as ideal, when in fact, this should be done by a specialist in phytopathology.

        \begin{table}[H]
            \centering
            \resizebox{0.3\textwidth}{!}{% <------ Don't forget this %
            \begin{tabular}{|c|c|}
                \hline
                Risk & Severity \\ \hline
                very low & up to 1\% \\ \hline
                low & between 1\% - 10\% \\ \hline
                middle & between 10\% - 20\% \\ \hline
                high & between 20\% - 40\% \\ \hline
                very high & between 40\% - 100\% \\ \hline
                \end{tabular}% <------ Don't forget this %
            }
            \caption{Risk of severity \cite{rastogi2015leaf}.}\label{tab:riscoSeve}
        \end{table}
        
\begin{figure}[H]
            \centering
            \subfloat[Input image.\label{fig:n1}]
            {\includegraphics[width=0.15\textwidth]
                {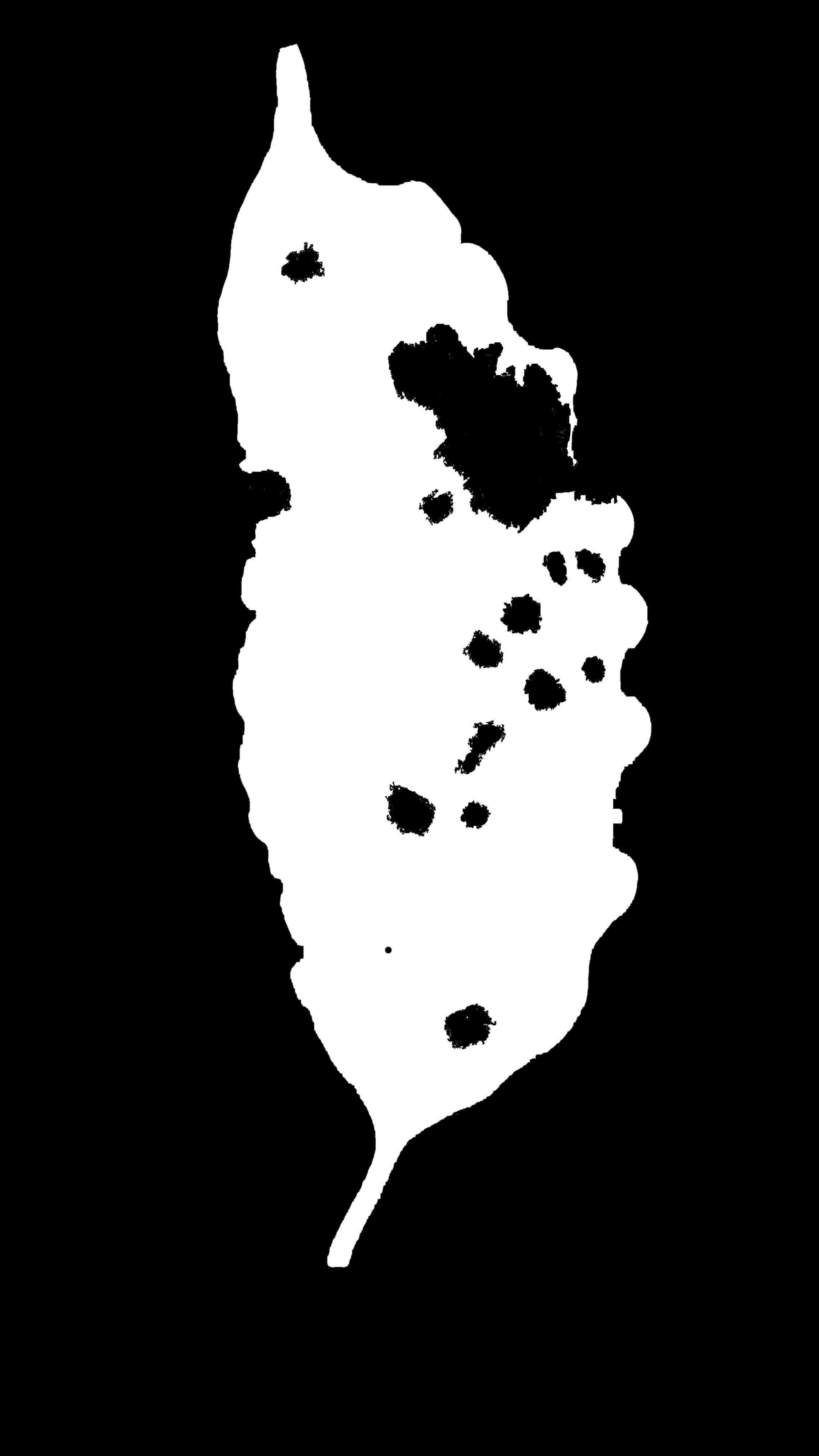}} \hspace{0.1cm}
            \subfloat[Obtained with YCgCr.\label{fig:n2}]
            {\includegraphics[width=0.15\textwidth]
                {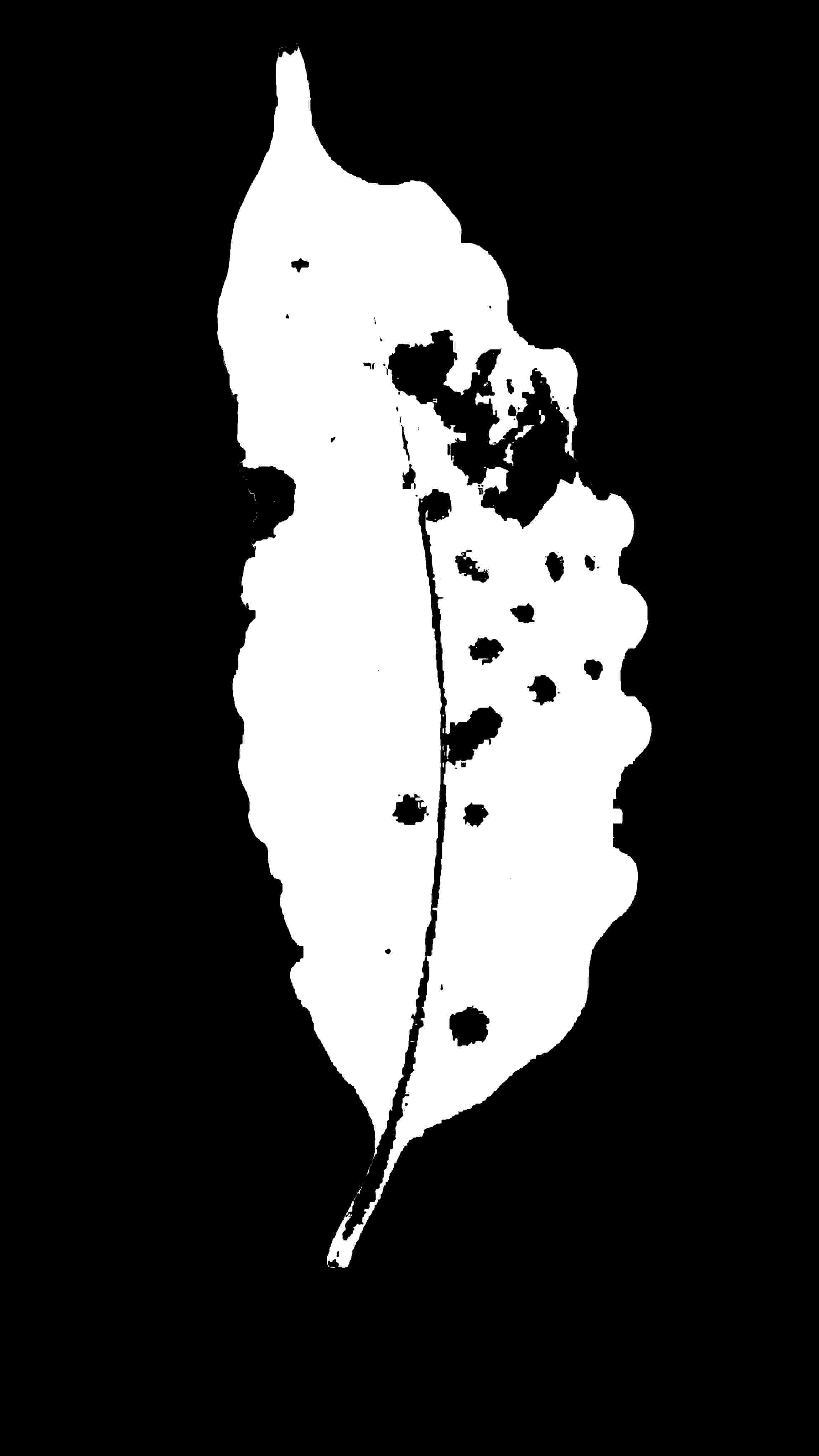}} \hspace{0.1cm}
            \subfloat[Result.\label{fig:n3}]
            {\includegraphics[width=0.15\textwidth]
                {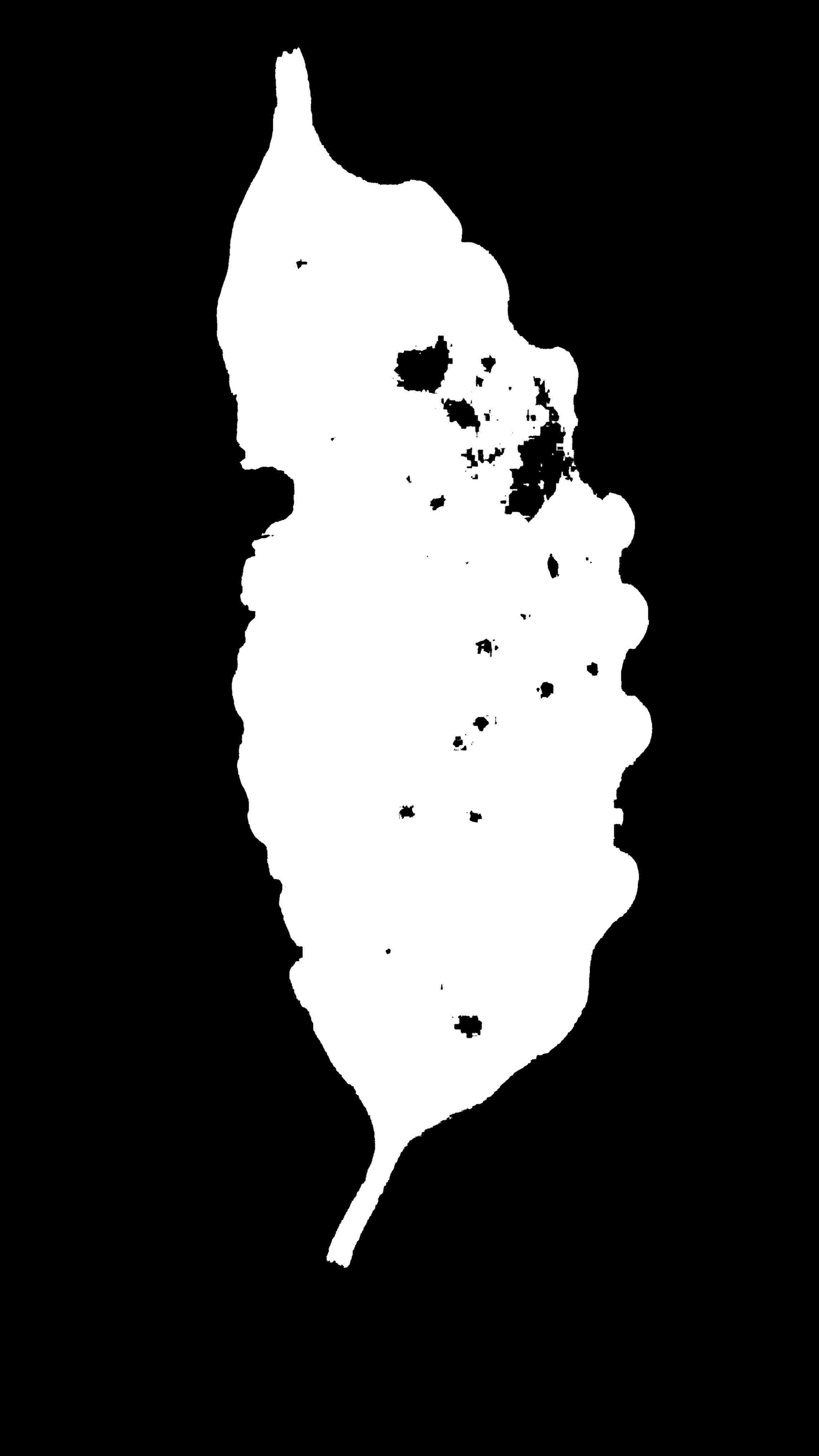}} \hspace{0.1cm}
            \caption{Comparison between the injured leaf segmentation masks for the 3 different methods.} 
            \label{fig:falhaSeveridade1}
        \end{figure}

        For image 5 of \autoref{tab:resultSeve}, the result of the severity using the segmentation in YCgCr color space  presented great error compared to the other two methods. This is because of the difference between manual segmentation and automatic segmentation in YCgCr. The masks generated by the different methods for this case is deployed in \autoref{fig:falhaSeveridade1}.
        
        %\begin{figure}[H]
        %    \centering
        %    \includegraphics[scale=0.3]{figs/27.png}
        %    \caption{\label{fig:falhaSeveridade1} Comparison between the lesion segmentation masks for the 3 different methods.}
        %\end{figure}

\begin{figure}[H]
            \centering
            \subfloat[Input image.\label{fig:o1}]
            {\includegraphics[angle=-90,width=0.15\textwidth]
                {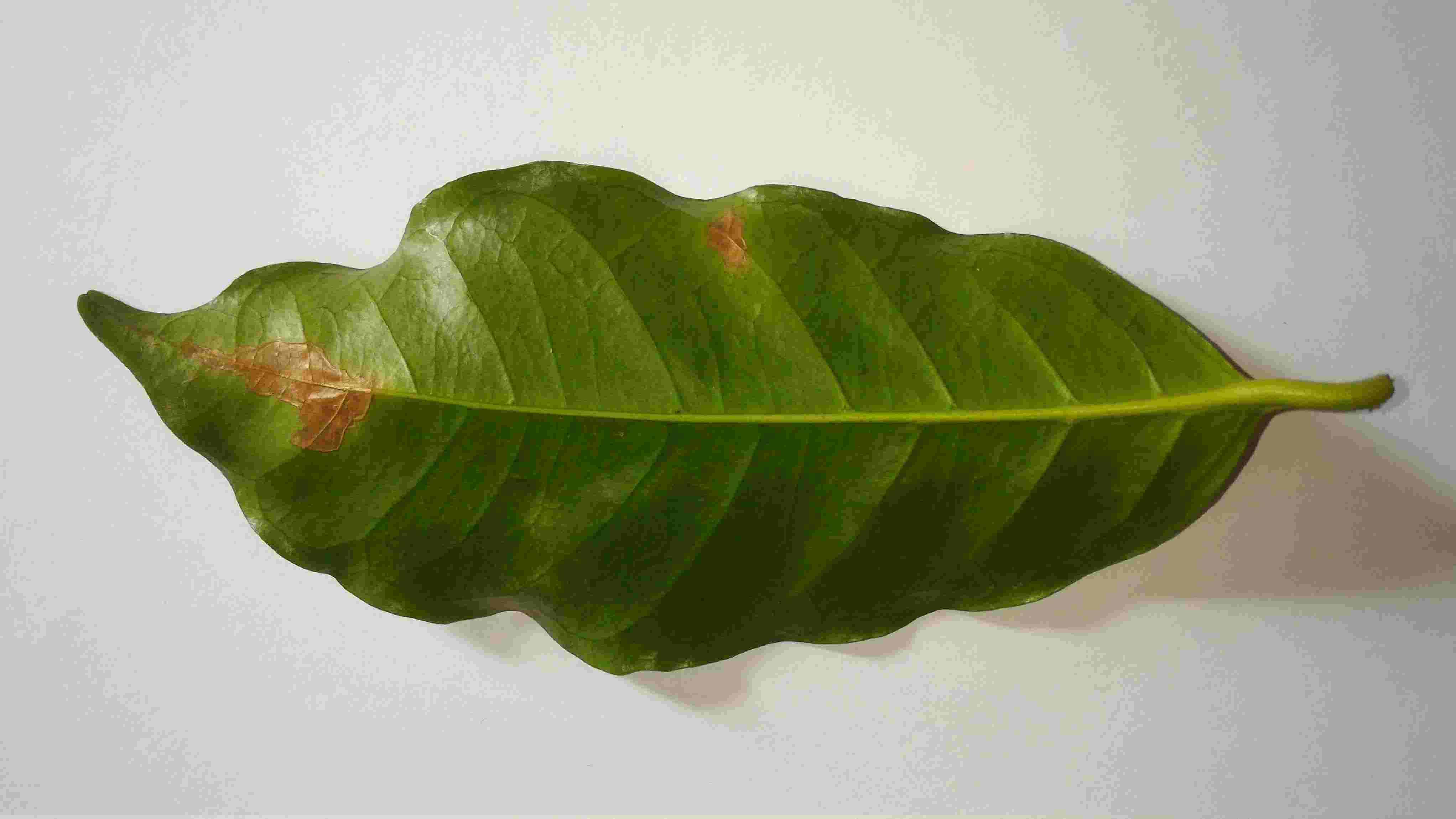}} \hspace{0.30cm}
            \subfloat[Segmented.\label{fig:o2}]
            {\includegraphics[angle=-90,width=0.15\textwidth]
                {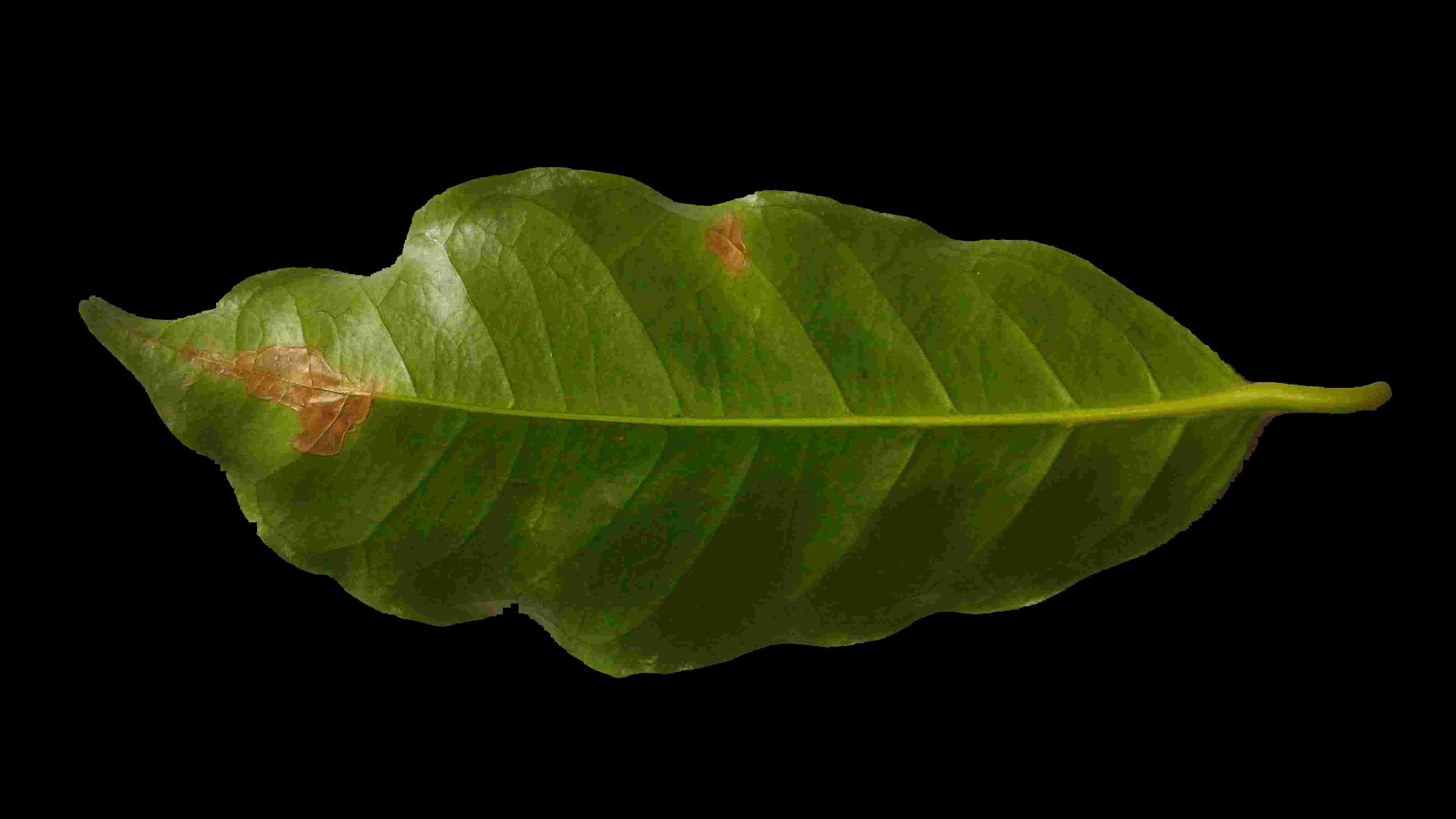}} \hspace{0.30cm}
            \subfloat[Result.\label{fig:o3}]
            {\includegraphics[angle=-90,width=0.15\textwidth]
                {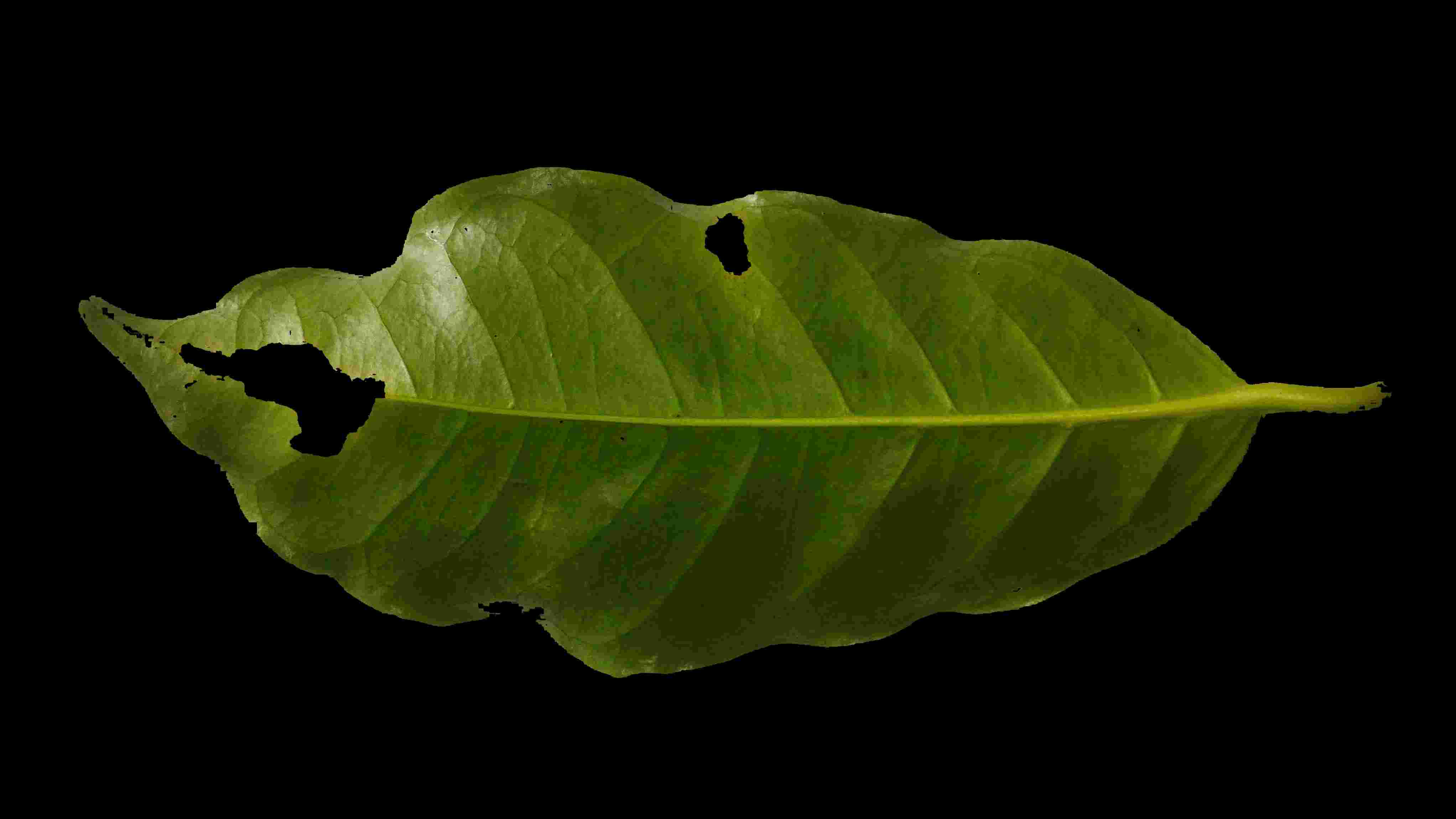}} \hspace{0.30cm}
            \caption{Pest - Coffee leaf miner; Severity: $3.62\%$.}
            \label{fig:severidade5}
        \end{figure}

\begin{figure}[H]
            \centering
            \subfloat[Input image.\label{fig:p1}]
            {\includegraphics[width=0.15\textwidth]
                {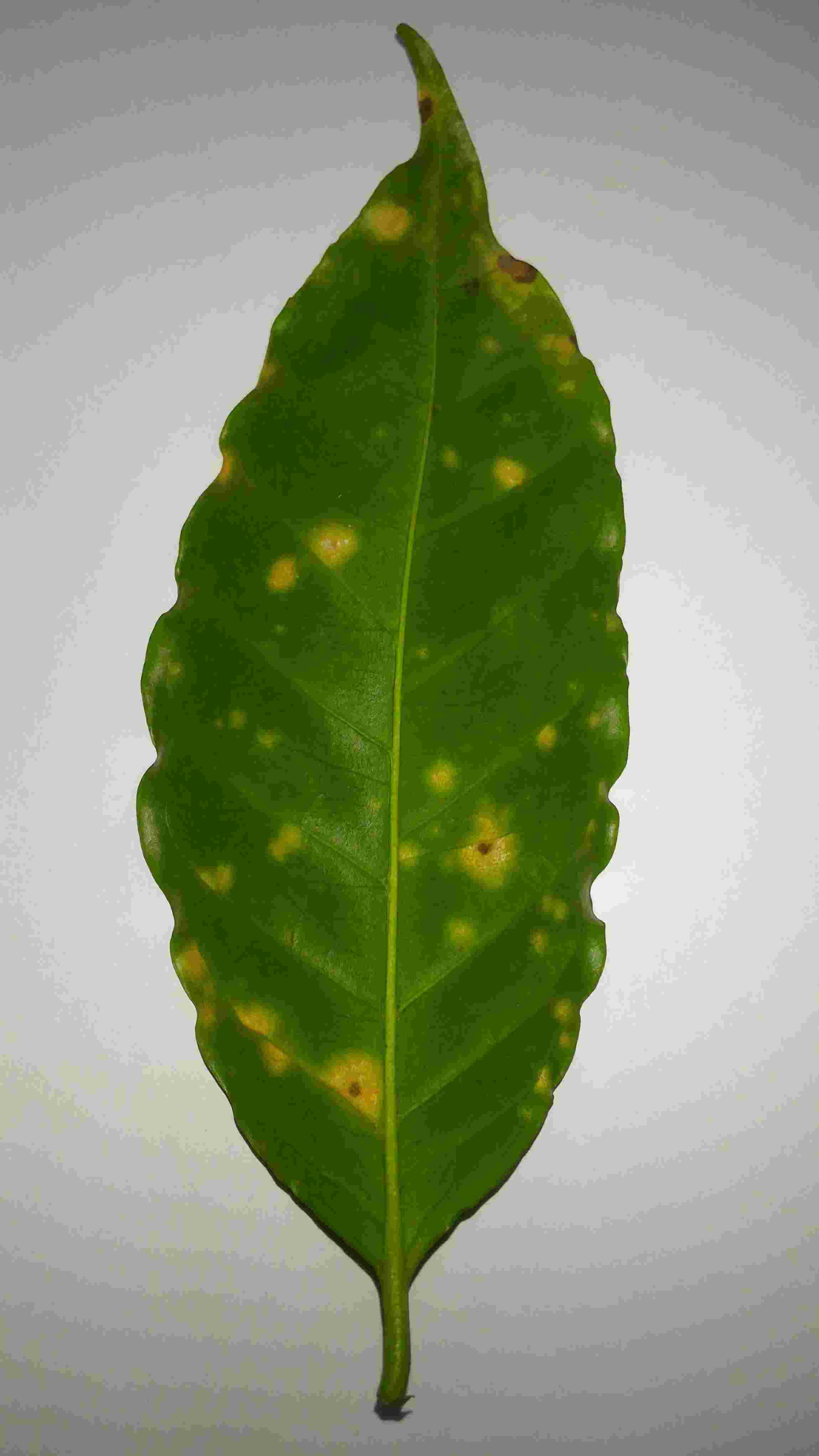}} \hspace{0.30cm}
            \subfloat[Segmented.\label{fig:p2}]
            {\includegraphics[width=0.15\textwidth]
                {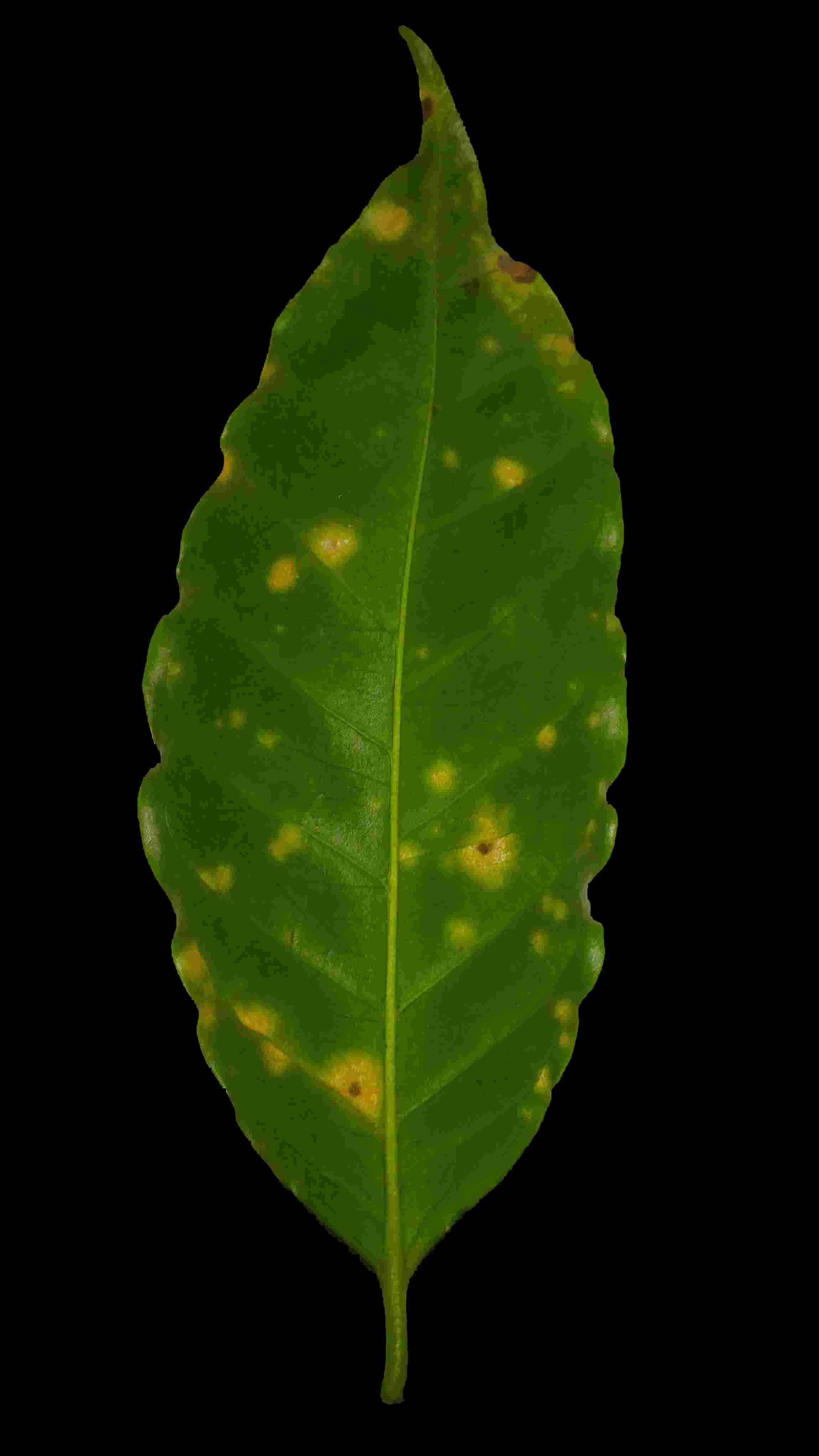}} \hspace{0.30cm}
            \subfloat[Result.\label{fig:p3}]
            {\includegraphics[width=0.15\textwidth]
                {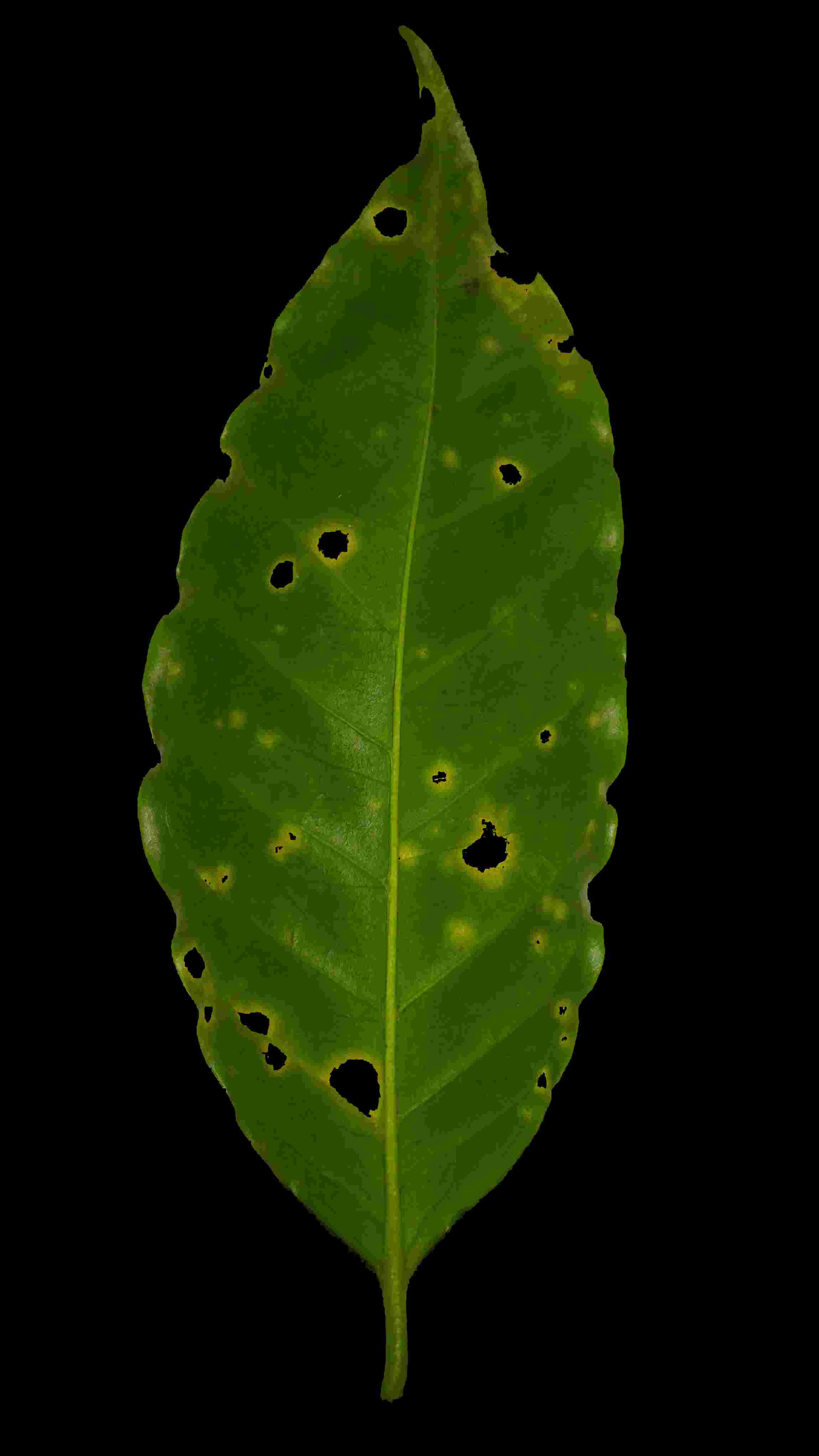}} \hspace{0.30cm}
            \caption{Disease - Coffee leaf rust; Severity: $2.33\%$.}
            \label{fig:severidade6}
        \end{figure}

        In fact, it is clear why the severity calculation is so different. The mask generated by the YCgCr method could not identify the entire damaged foliar area of the leaf. Figures \ref{fig:severidade5} and \ref{fig:severidade6} show the images after the processing with the classification of the injured area and the calculation of severity.
        
        %\begin{figure}[H]
         %   \centering
         %   \includegraphics[scale=0.3]{figs/28.png}
         %   \caption{\label{fig:severidade5} Pest: Leaf Miner; Severity: $3.62\%$.}
        %\end{figure}

        %\begin{figure}[H]
        %    \centering
        %    \includegraphics[scale=0.3]{figs/29.png}
        %    \caption{\label{fig:severidade6} Disease: Rust; Severity: $2.33\%$.}
        %\end{figure}

\section{Development of an Application (App)}

\begin{figure}[H]
        \centering
        \includegraphics[scale=0.8]{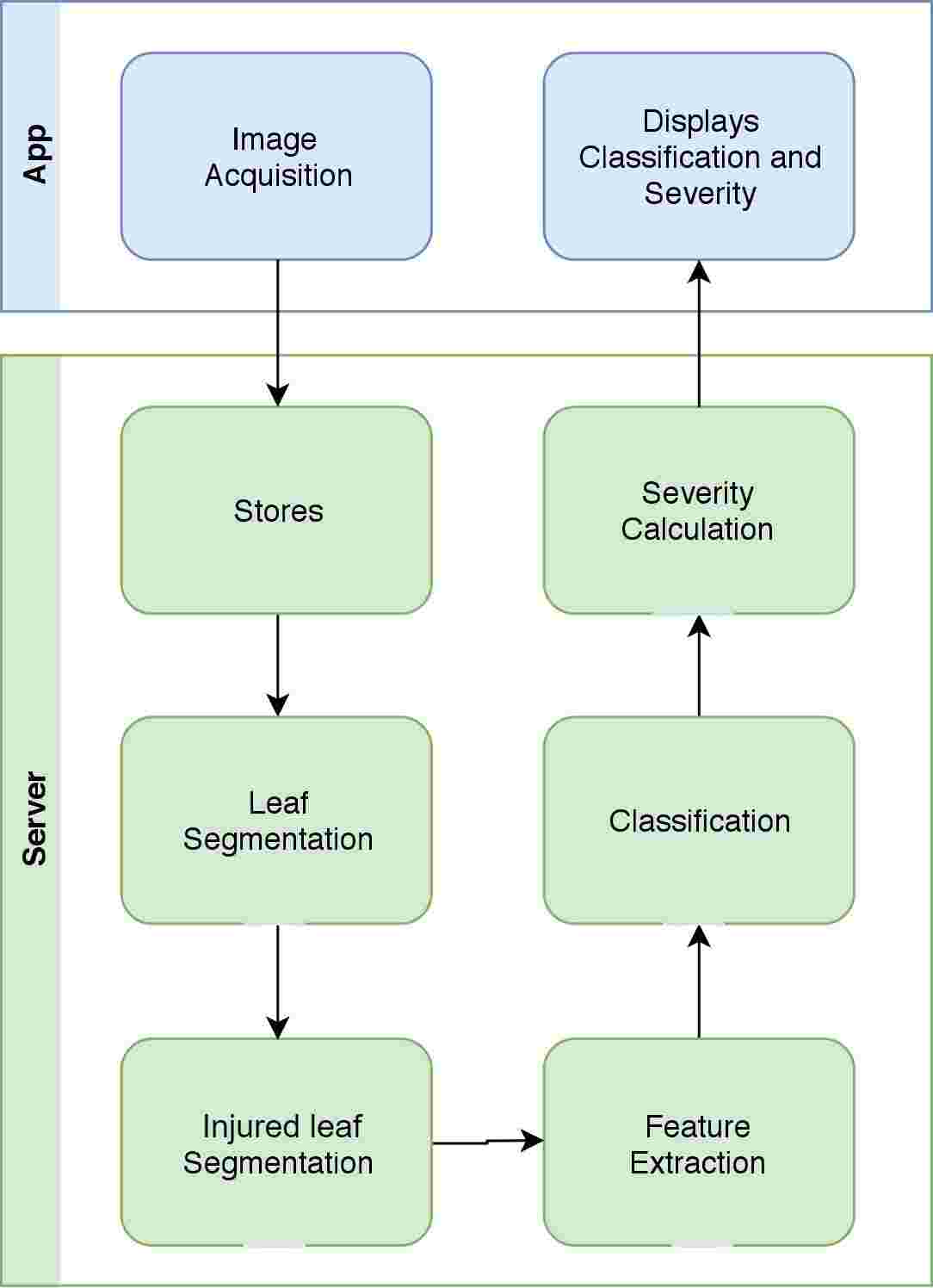}
        \caption{\label{fig:platVisal} Classification process and severity calculation.}
    \end{figure}

    With the ease of access to the internet and devices such as smartphones, the number of applications that aim to facilitate the tasks for the final users has increased. So, the application developed in this work seeks to integrate the whole process of segmentation of the coffee leaf and its damaged foliar area deploying the results of classification and severity. For the development of the application, the Android system was chosen as it is widely used and has an affordable development platform. The entire image processing step was performed on a server, which receives the image captured by the smartphone, stores it  and generates a document with the information to the database, performs the necessary processing by executing the code in Python, saves the information resulting from the processing in the database and sends the results to the application. Full-Stack Mean.js was used for the server development. The platform overview is shown in \autoref{fig:platVisal}.
    
    \subsection{Used Tecnologies}
    
        \subsubsection{Framework Mean.Js}
        
            The acronym was first used in 2013 by Karpov's development team of MongoDB to denote the use of a complete stack for development applications using MongoDB, Express, AngularJS and Node.Js \cite{chodorow2010mongodb}.
            
            \begin{itemize}
                \item \textbf{MongoDB:} It is a powerful and flexible NoSQL database. It combines the ability to scale with the many resources available in relational databases such as indexes, ordering, etc \cite{chodorow2010mongodb}. In the mean Stack, MongoDB allows you to store and retrieve a data in a format very similar to JavaScript Object Notation (JSON). 
                
                \item \textbf{Express:} It is a light-weight web framework, which helps in organizing your web application in MCV architecture on the server side. Express generally provides REST Endpoints that are consumed by the AngularJS templates, which in turn are updated based on the  received data \cite{almeida2015mean}.
                
                \item \textbf{AngularJs:} It is a client-side framework for Single Page Web Applications (SPA) created by Google and released to the public domain in 2009. It greatly simplifies the development, maintenance and testing, as well as facilitating the receipt of data and execution of logic directly on the client. In Mean Stack, it is responsible for dynamic interfaces allowing easy access to REST endpoints directly from the client through specialized services \cite{almeida2015mean}.
                
                \item \textbf{Node.Js:} It is a platform for JavaScript applications created by Dahl under the Chrome JavaScript runtime environment. It is responsible for server-side requests \cite{cantelon2014node}.
            \end{itemize}
            
        \subsubsection{Android System}
            
            Android is the most widely used operating system on smartphones and tablets. It is based on the Linux kernel and currently developed by Google. Android Studio and the Java language was used for the development of the application.  Regarding the communication with the Android system the Software Development Kit (SDK) is used, in which it allows the developer to use the functionalities provided by the operating system. 
        
    \subsection{Functionalities}
    
\begin{figure}[H]
            \centering
            \includegraphics[scale=0.4]{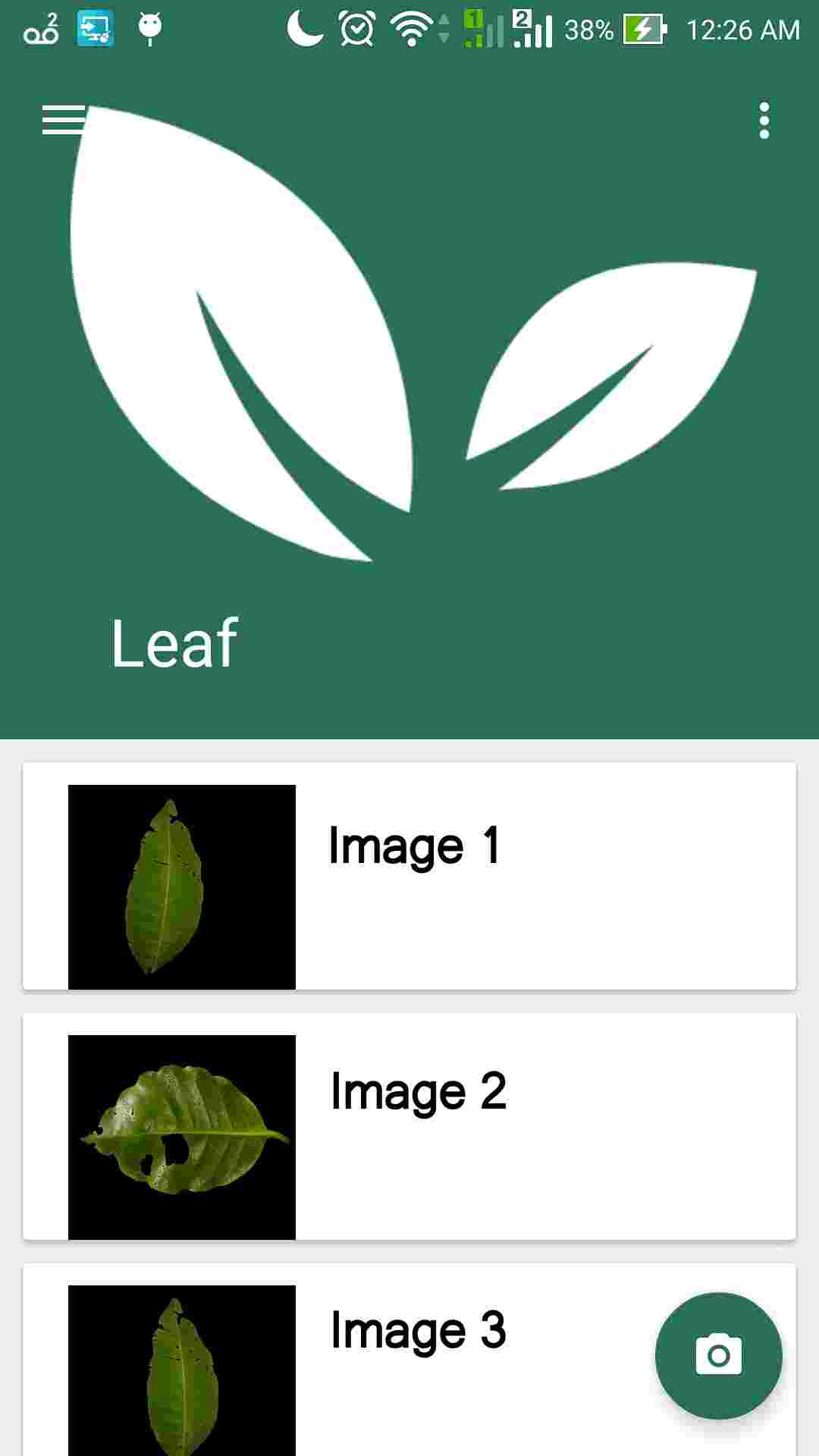}
            \caption{\label{fig:tela2} Initial screen with the list of previous classifications.}
        \end{figure}

        After the user log in, the app pops up the initial screen with the list of all the evaluations already performed as deployed in \autoref{fig:tela2}. 
        
\begin{figure}[H]
            \centering
            \includegraphics[scale=0.4]{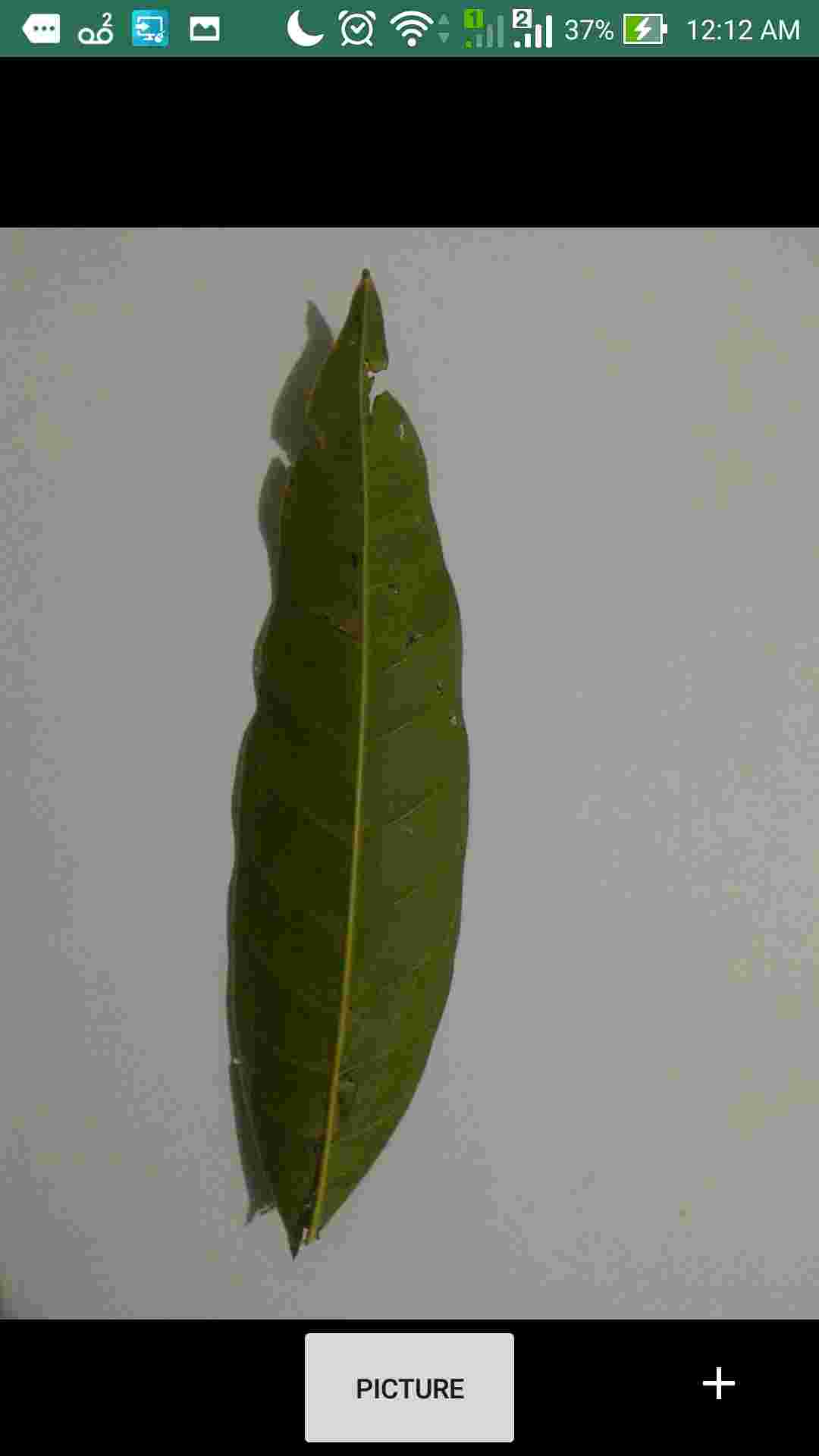}
            \caption{\label{fig:tela3} Screen showing a coffee leaf to be selected.}
        \end{figure}

        To re-evaluate, the user clicks on the camera icon of the \autoref{fig:tela2} and then he/she is redirected to the screen deployed in the \autoref{fig:tela3}.
        
\begin{figure}[H]
            \centering
            \includegraphics[scale=0.4]{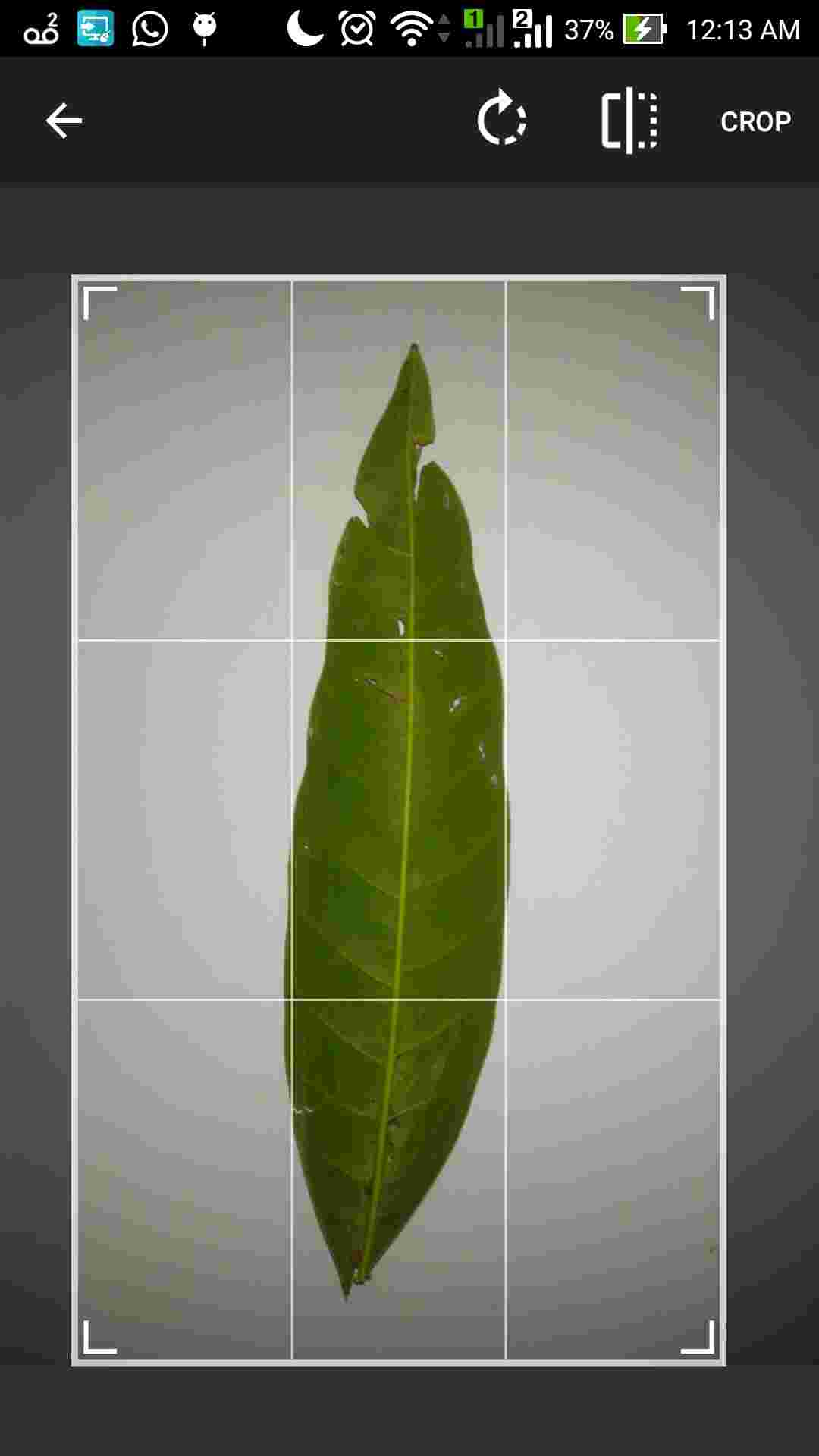}
            \caption{\label{fig:tela4} Screen to select the coffee leaf.}
        \end{figure}

        After capturing the image, the user must select the region of interest within a grid, so that the leaf is highlighted as shown in \autoref{fig:tela4}.

\begin{figure}[H]
            \centering
            \includegraphics[scale=0.4]{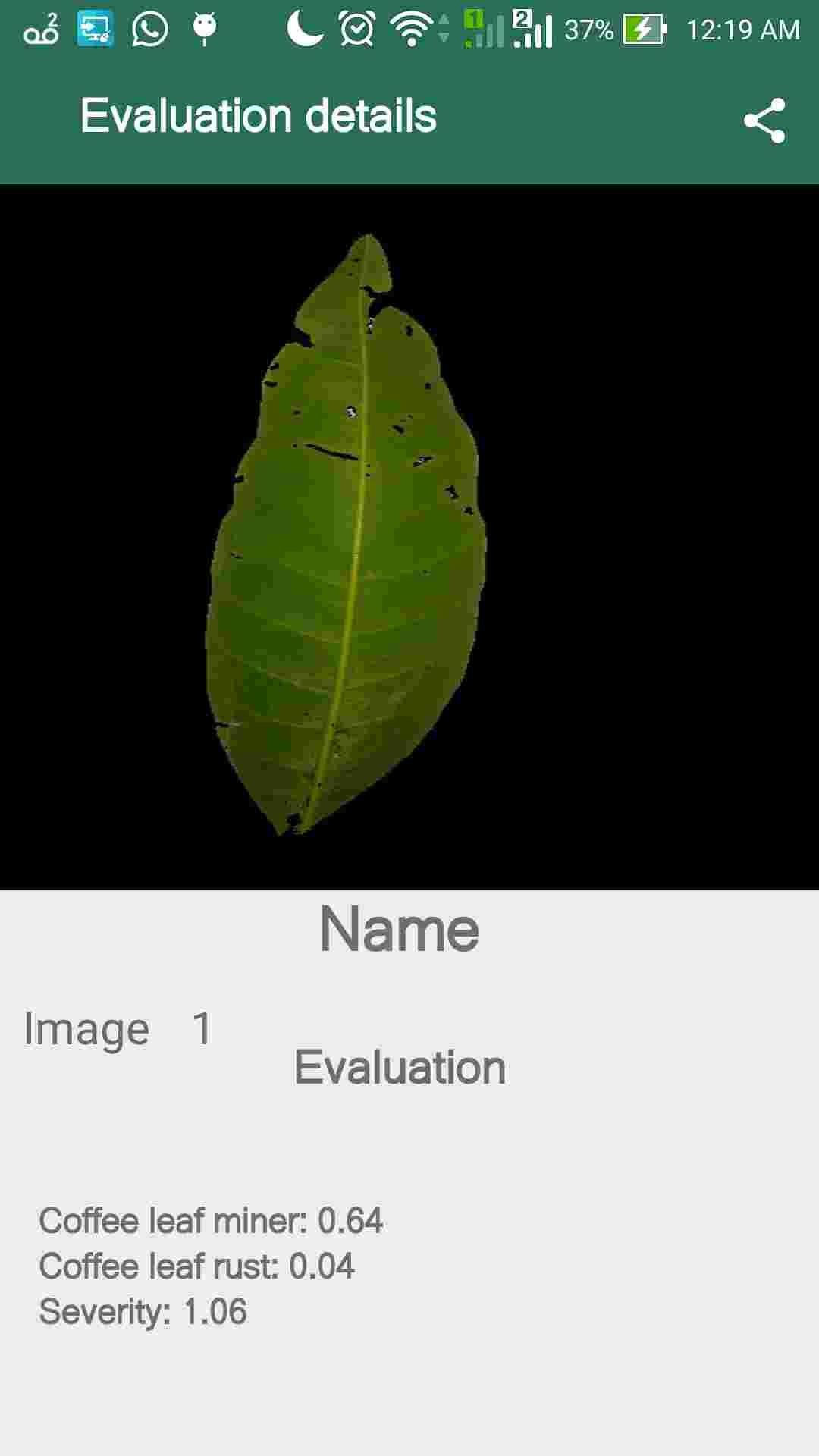}
            \caption{\label{fig:tela5} Screen with details of the automatic classification and severity.}
        \end{figure}

        After cropping, the image may be uploaded and processed on the server, and the automatic evaluation can be viewed as shown in \autoref{fig:tela5}.
        
\section{Conclusion}

    In this work, we presented the development of a system (an App) that automatically determine the type of coffee leaf disease/pest and the percentage of the injured area. In each process of the development, different approaches have been compared in order to find out the best methods for each case. In image segmentation, the \textit{k-means} algorithm, the Otsu method and the iterative threshold method have been tested on YCgCr color space.  For feature extraction attributes of texture and color were calculated. For classification, Artificial Neural Network trained with Backpropagation and Extreme Learning Machine were used. The presented methodology allows an automatic process of classification and calculation of severity of the damaged area of coffee leaves. According to the obtained  results in the segmentation, it was clear the importance of choosing the right background to obtain good results. For each chosen background there is a component in the HSV or YCbCr colors space that produce the best result according to the metric used. In general, there is no  method or configuration that works well in all cases. In the process of segmentation of damaged foliar area, it was possible to observe more clearly the influence of the background on the leaf color. The camera's automatic adjustment, which depends on the ambient colors, generated images with different shades of green. This had a direct impact on the segmentation of damaged foliar areas of the leaf, which in some images with black or blue background, have generated bad results. For this reason, in this article, the white background is indicated. In the classification process using Artificial Neural Network trained with Backpropagation and Extreme Learning Machine (ELM) good results have been obtained according to the metrics used, but ELM was slightly better. Although the damaged foliar areas of the leaf are noticeably different in color and texture, the good results in the classification show that the attributes chosen were suitable for the this purpose.
    
    The  automatic calculation of the severity proposed  in this article was compared with the manually using Photoshop by means of the calculation of the injured area and with  evaluation of experts and the results are very close  showing the suitability of the approach. Small variations in severity are tolerable in practice, since severity is evaluated in ranges of values. However, the use of backgrounds other than white may generate non consistent results. The next steps for future work is to investigate segmentation methods that are more robust, i.e., that present good results for various types of backgrounds. In addition, we suggest the development of new metrics to evaluate the segmentations and the expansion of the dataset by acquiring more images. As we know there are other diseases and pests that affect coffee leaves. In this case, a deeper study should be made regarding the features extraction and the classifier model. Another point is that with some  alterations, the method may be adapted to other plants.

\section{Acknowledgements}

We gratefully acknowledge the support of NVIDIA Corporation with the donation of the Titan Xp GPU used for this research R.A. Krohling would like to thank the Brazilian agency CNPq and the local Agency of the state of Espirito Santo FAPES for financial support under grant No. 309161/2015-0 and No. 039/2016, respectively.

%\section{Bibliography styles}

%There are various bibliography styles available. You can select the style of your choice in the preamble of this document. These styles are Elsevier styles based on standard styles like Harvard and Vancouver. Please use Bib\TeX\ to generate your bibliography and include DOIs whenever available.

%Here are two sample references: \cite{Feynman1963118,Dirac1953888}.

\section{Conflict of Interest Statement}
The authors declare that they have no conflicts of interest.

\section*{References}

\bibliography{mybibfile}

\end{document}